\documentclass{article}

    \PassOptionsToPackage{numbers, compress}{natbib}
 \usepackage[preprint]{neurips_2026}

\usepackage[utf8]{inputenc} % allow utf-8 input
\usepackage[T1]{fontenc}    % use 8-bit T1 fonts
\usepackage{hyperref}       % hyperlinks
\usepackage{url}            % simple URL typesetting
\usepackage{booktabs}       % professional-quality tables
\usepackage{amsfonts}       % blackboard math symbols
\usepackage{nicefrac}       % compact symbols for 1/2, etc.
\usepackage{microtype}      % microtypography

\usepackage{booktabs}
\usepackage{multirow}
\usepackage{graphicx}
\usepackage[table,xcdraw]{xcolor}
\usepackage[normalem]{ulem}
\useunder{\uline}{\ul}{}
\usepackage{makecell, multirow}
\usepackage{pifont}

\newcommand{\xmark}{\ding{55}}
\usepackage{wrapfig}
\usepackage{caption}
\usepackage{tabularx}
\usepackage[table]{xcolor}
\usepackage{float}

\title{HAPACT: A Benchmark For Human-Centric \\ Physical Impact Localization in Movies}
\author{%
  Youngrae Kim\textsuperscript{2}, Yejin Jang\textsuperscript{3}, Eunho Kim\textsuperscript{1}, Youjin Sung\textsuperscript{1},\\
  \textbf{Kun-Woo Song\textsuperscript{1}, Sangpil Kim\textsuperscript{4}, Sang Ho Yoon\textsuperscript{1,2,3}}\\
  \\
  \textsuperscript{1} Graduate School of Culture Technology, KAIST, South Korea\\
  \textsuperscript{2} Graduate School of Metaverse, KAIST, South Korea\\
  \textsuperscript{3} Department of AX, KAIST College of AI, South Korea\\
  \textsuperscript{4} Department of Artificial Intelligence, Korea University, South Korea\\
}

\begin{document}

\maketitle

% keywords: Spatio-Temporal Video Grounding, Human Physical Understanding, Human-Centric Video Understanding, Video-to-Haptic Generation.

\begin{abstract}

Physical reasoning about humans in videos has largely focused on everyday interactions, where contact, pressure, pose, and motion support embodied perception and contact-aware applications. Still, short and forceful properties like impact events remain relatively underexplored, despite their relevance to broader understanding of how abrupt physical events affect the human body. To fill this gap, we introduce human-centric impact localization, a new task that localizes target physical impacts temporally within a video and spatially on the human body. We further present \textbf{HAPACT}, a \underline{H}um\underline{A}n-centric im\underline{PACT} dataset and benchmark built from movie clips, containing over $2.8$K video-query pairs with more than $20$K fine-grained human-annotated frames. We built HAPACT around viewer-grounded annotation, text-conditioned localization, and supervision at frame-level temporal and SMPL-X-based spatial granularity. To demonstrate the effectiveness of our benchmark, we conduct extensive evaluations of existing methods, employing video moment retrieval models for temporal localization and human-scene contact models for spatial localization. For the spatial task, we propose a text-conditioned baseline by reformulating localization as query-guided body-region selection. Benchmark results show that physical impact localization remains challenging for existing methods, motivating HAPACT as a dedicated benchmark to support physical impact reasoning in videos.

\end{abstract}

\section{Introduction}

Physical AI and multimodal foundation models increasingly require human-centered physical reasoning from videos, going beyond action recognition to understand how physical events affect the human body. Recent body-centric physical interaction datasets examine how the body interacts with surrounding scenes~\citep{chen2023detecting, hassan2019resolving, huang2022capturing, tripathi2023deco}, objects~\citep{cseke2025pico,  jiang2023full, lu2025humoto, taheri2020grab, xie2022chore, yang2024egochoir}, and other people~\citep{khirodkar2024harmony4d, xu2024inter, yin2023hi4d}. They mainly focus on everyday user-driven interaction regimes, providing supervision for physically grounded cues such as contact, pressure, pose, and motion. Still, externally initiated interactions such as bodily impacts have remained overlooked.

Impact events offer a complementary axis for human-centric physical understanding. Unlike sustained or continuous interactions, impacts are short and forceful events in which the human body is suddenly subjected to external force~\citep{gilardi2002literature}. Understanding such events requires reasoning about when and how external elements affect the body, as well as where those effects occur. Unlike common physical cues that arise from user-initiated interactions, impacts often arise from external agents or the environment, making them difficult to capture with active interaction cues alone. This shifts the focus from sustained interaction modeling to a broader understanding of how both user-initiated and externally initiated physical events affect the human body.

Action-oriented movies provide a rich domain for understanding impact, as they frequently depict bodily impacts by direct collisions and environmental forces in realistic scenes. Beyond their diverse impact depictions, movies also offer rich scenes and narrative context for video understanding, as leveraged in prior movie-based benchmarks~\citep{gu2018ava, song2024moviechat}. This context requires models to identify the specific impact described by a text query, rather than simply detect any events in the video. Reasoning about physical events within their video context can also support applications in immersive media such as haptic authoring and automated haptic experience generation~\citep{john2026hapticlens, jung2024hapmotion, li2021towards}.

This motivates our formulation of \textbf{Human-Centric Impact Localization}, a new task that localizes a target physical impact temporally within a video and spatially on the human body. Building on this task, we introduce \textbf{HAPACT}, a \underline{H}um\underline{A}n-centric im\underline{PACT} dataset and benchmark for impact localization in movie clips. The dataset contains over $2.8$K video-query pairs with more than $20$K fine-grained human-annotated frames. To make the annotation procedure tractable, HAPACT uses a semi-automated pipeline that first generates coarse candidates with a vision-language model, then annotators create fine-grained labels using an interactive web-based interface. HAPACT is designed around three key properties:

\begin{figure*}[t]
  \centering
  \includegraphics[width=\textwidth]{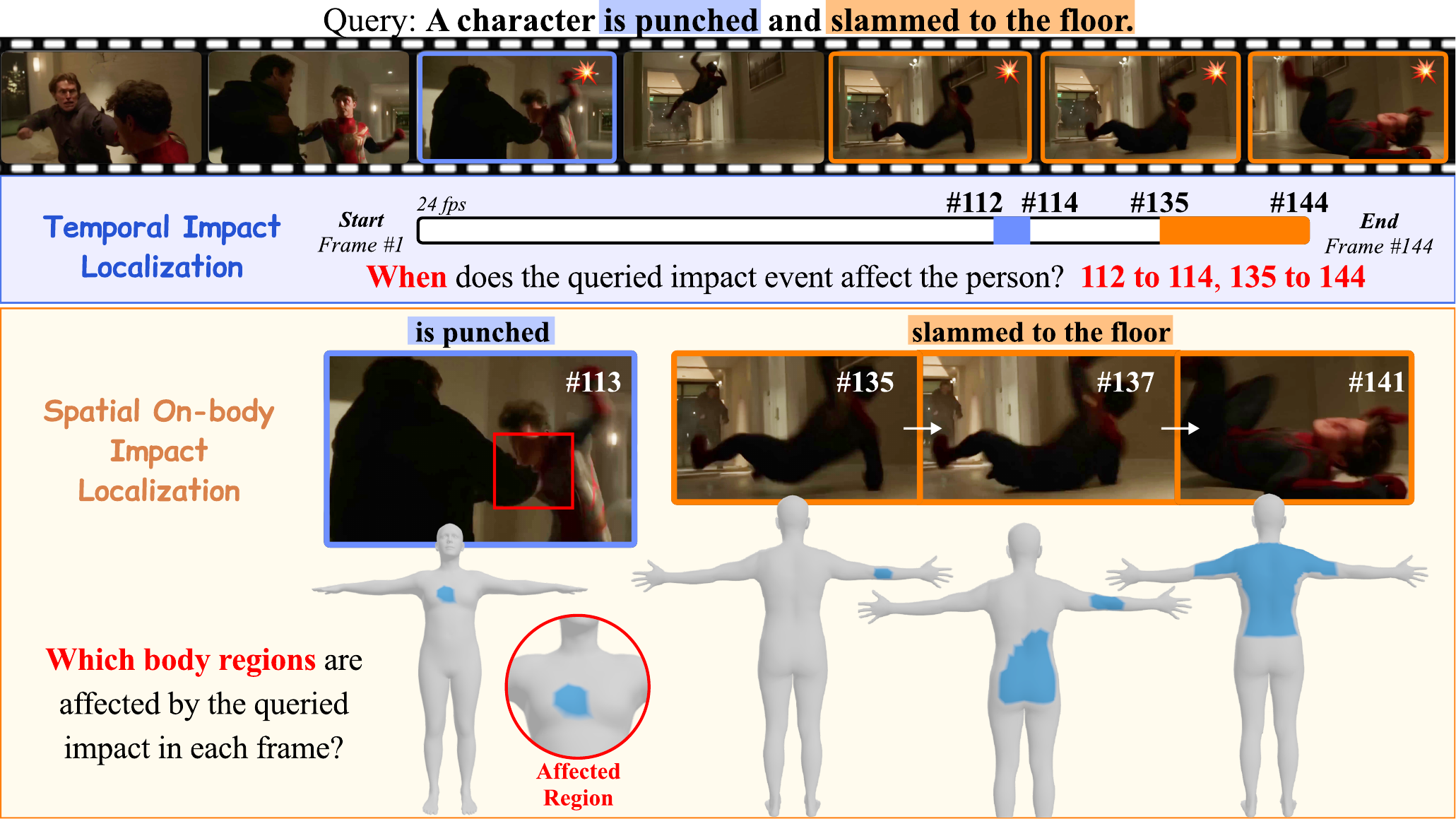}
  \caption{\textbf{Human-Centric Impact Localization.}
    Given a video and a text query describing a target impact event, the task requires localizing when the queried impact affects the person and which body regions are affected in each frame.}
  \label{fig:task_definition}
\end{figure*}

\begin{itemize}

    \item \textbf{Viewer-grounded annotation.}
    All annotations are labeled for impacts identified by human viewers. This allows HAPACT to better align its annotations with natural human interpretation of impact events in the video context and cinematic cues.

    \item \textbf{Text-conditioned localization.}
    Each textual query specifies a target impact event, and the task requires localizing the target event rather than the whole impact in the scene. This query-conditioned formulation supports event-level reasoning in complex movie scenes.

    \item \textbf{Fine-grained supervision.}
    HAPACT provides frame-level temporal annotations at 24~FPS and SMPL-X~\citep{pavlakos2019expressive} mesh annotations for fine-grained spatio-temporal evaluation.
    
\end{itemize}

As a benchmark, HAPACT enables multi-axis evaluation across temporal and spatial axes. We evaluate temporal localization with existing video moment retrieval models and spatial localization with human-scene contact models. Since existing human-scene contact models are not conditioned on textual input, we further formulate on-body impact localization as query-guided body region selection and instantiate it by fine-tuning a multi-modal large language model (MLLM). While this region selection baseline outperforms contact-based models~\citep{huang2022capturing, lin2026graphicontact, tripathi2023deco}, it remains limited to coarse body regions rather than fine-grained estimations on the body mesh. By bringing bodily impacts into physical reasoning, HAPACT broadens human physical understanding beyond user-intended interaction, providing a benchmark foundation for richer reasoning about dynamic environmental events and their consequences.

To summarize, the contributions are as follows:

\begin{itemize}

    \item We formulate human-centric impact localization, a new task that identifies each queried physical impact with its corresponding video frames and predicts the affected region of the human body.

    \item HAPACT is the first human-centric video benchmark for physical impact events, providing viewer-grounded, text-conditioned annotations with frame-level SMPL-X labels.

    \item We evaluate representative models for temporal localization and spatial on-body localization, employing video moment retrieval models and human-scene contact models. We further introduce an MLLM baseline for query-guided region selection.

\end{itemize}

\section{Related Work}

\paragraph{Body-centric Physical Interaction Datasets.} Recent body-centric datasets have expanded from modeling humans in physical scenes to richer cues about how the body interacts with its surroundings. \citet{hassan2019resolving} was an early effort to use contact and penetration constraints to recover physically plausible human poses in real environments, while subsequent studies~\citep{chen2023detecting, huang2022capturing, shinbodycontact4d, tripathi2023deco} extended this direction toward dense body representations, including SMPL-X~\citep{pavlakos2019expressive}, and contact estimation in in-the-wild images. Beyond scene constraints, later datasets broaden this perspective to other interaction domains, spanning from human-object interaction~\citep{bhatnagar2022behave, cseke2025pico, huang2022intercap, lu2025humoto, taheri2020grab, xie2022chore, yang2024lemon, yang2024egochoir} to human-human interaction~\citep{khirodkar2024harmony4d, xu2024inter, yin2023hi4d} between multiple bodies. Recent datasets further enrich body-centric physical cues like hand-pressure estimation, but remain largely focused on hand interactions~\citep{grady2022pressurevision, seo2024posture, zhao2025egopressure}. While these datasets provide rich supervision for body-centric physical cues, they primarily cover everyday or user-initiated interactions, leaving external, environment-initiated physical events less represented. HAPACT complements this line by introducing impact as a physical cue in movie videos, capturing bodily effects from short, abrupt external force.

\paragraph{Video Moment Retrieval and Spatio-temporal Grounding Benchmarks.}
Video localization benchmarks~\citep{anne2017localizing, gao2017tall, krishna2017dense, lei2021detecting, regneri2013grounding, Zala2023HiREST} have studied temporal event understanding not only through action localization, but also through moment retrieval guided by language queries across diverse video domains. Recent benchmarks further expand this language-guided moment retrieval along several axes, including corpus-level retrieval~\citep{chen2024verified, lei2020tvr}, egocentric~\citep{grauman2022ego4d, perrett2025hdepic}, long~\citep{Croitoru_2023_ICCV, soldan2022mad, yuan2025momentseeker} video domains, and multiple relevant moments~\citep{cao2025flashmmr, ICLR2025_7ac19fdc}. 

Beyond temporal localization, spatio-temporal video grounding~(STVG) localizes targets specified by language jointly across video frames and image space. Existing STVG benchmarks~\citep{9446308, zhang2020does} provide foundations for joint spatial and temporal localization. Here, their spatial targets are usually defined in image coordinates, using boxes or tubes around target instances. HAPACT follows this formulation but shifts the spatial target from image-level anchors to frame-level body regions. This representation captures the effect of a queried physical event on the human body.

\section{HAPACT}
\label{sec:dataset}

HAPACT is a movie-based dataset for human-centric impact understanding in videos. We target short and forceful events in which the human body is affected by collisions, falls, object interactions, or environmental forces. The dataset is built from over $5$ hours of high-resolution videos spanning $212$ movies, and contains over $2.8$K video-query pairs with clips ranging from $6$ to $8$ seconds. The following sections detail the data collection and annotation procedures.

\subsection{Data Collection and Preprocessing}

\paragraph{Video Collection and Clip Preprocessing.}
Following~\citet{bain2020condensed}, we constructed HAPACT from movie clips collected from \textit{Movieclips}\footnote{\url{https://www.youtube.com/@movieclips}}, an official, publicly accessible YouTube channel with stable long-term availability. Movie titles are parsed from video titles and descriptions, and metadata~(e.g., release year and genre) are retrieved from \citet{imdb}. Non-film works, such as TV shows and documentaries, and adult-rated movies are excluded from data collection. 

\paragraph{Genre-based Filtering.}
Since impact events are not evenly distributed across movie genres, we applied genre-based filtering to prioritize clips more likely to contain salient impact scenes. A genre prior was constructed from films screened in 4DX~\citep{cj4dx} format, by computing both genre frequencies and genre-pair frequencies. We assigned a genre-prior score to each candidate movie based on its individual genres and genre pairs, then retained movies above the threshold. The detailed scoring formulation and procedure are described in Section~\ref{sec:genre-based scoring}.

\paragraph{Query-clip Candidate Generation via Vision-Language Model.}
To generate candidate instances, all movie clips were processed with Gemini 2.5 Pro~\citep{comanici2025gemini} to identify impact-related moments and produce second-level temporal intervals with textual descriptions of the corresponding events. For each candidate event, a segment was cropped around the estimated interval with a $\pm 2.5$~second buffer, preserving contextual cues before and after the impact. Candidate segments longer than $8$~seconds were excluded to focus on short, salient impact events.

\paragraph{Human-targeted Impact Filtering and Manual Revision.} 
Using the generated textual descriptions, human-targeted impacts were identified through two parallel filters. GPT-4o~\citep{hurst2024gpt} was prompted to identify descriptions in which the impact involves a human, while a Stanford Parser-based filter, inspired by the human-subject filtering of \citet{gupta2015visual}, retained descriptions whose affected argument denotes a person. The outputs of the two filters were merged to form the candidate pool. The candidates were then manually reviewed and revised by members of the research team to remove cases outside the task scope. To assess whether the candidate pipeline captures human-targeted impacts in the videos, we further evaluated its recall, obtaining a mean recall of 88.99\%. Detailed criteria and results are provided in Section~\ref{sec:construction_details}.

\begin{figure*}[t]
  \centering
  \includegraphics[width=\textwidth]{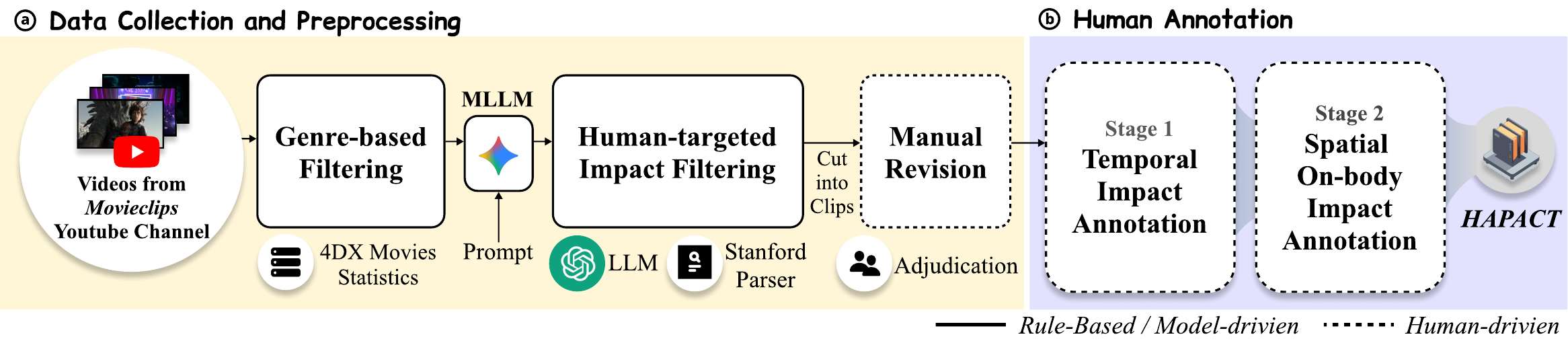}
  \caption{\textbf{Overview of the HAPACT Construction Pipeline.} Videos from the \textit{Movieclips} YouTube channel are filtered by genre priors, processed with Gemini 2.5 Pro~\citep{comanici2025gemini} to identify impacts and generate captions. Then, we refined videos through human-targeted filtering and manual revision. The clips are labeled in two phases, including temporal localization and spatial on-body regions.}
  \label{fig:construction_pipeline}
\end{figure*}

\subsection{Data Annotation}

HAPACT was annotated in two phases: temporal impact annotation and spatial on-body impact annotation. We used the HumanSignal~\citep{labelstudio2025} platform to conduct both phases with $28$ trained annotators, of whom $50\%$ were female. The annotator pool consisted of undergraduate and graduate students across diverse academic disciplines. The process took approximately $30$~weeks in total.

\paragraph{Stage 1: Temporal Impact Annotation.}
In the first phase, annotators labeled the frame-level temporal extent of the queried impact event for each video-query pair. All clips were re-encoded at $24$~FPS, and annotators were provided with frame-level access to each clip. The task required marking every occurrence of the queried impact, including repeated instances within the same clip. We collected three independent annotations for each pair and aggregated by majority vote to determine the final temporal intervals. Ambiguous cases after aggregation were resolved through manual adjudication. Section~\ref{sec:temporal_annotation_aggregation} provides the detailed procedure.

\paragraph{Stage 2: Spatial On-body Impact Annotation.}
In the second stage, annotators labeled the body region affected by the queried impact over time. For each temporally localized impact, annotators viewed the corresponding video segment and marked the affected region on an SMPL-X~\citep{pavlakos2019expressive} body mesh for each frame. To support this process, we developed a custom painting interface\footnote{A demo of the annotation interface is available at \url{https://smpl-painter.netlify.app/}.} inspired by prior dense 3D human annotation datasets~\citep{tripathi2023deco, yang2024lemon} and deployed it as a web-based tool for fine-grained on-body annotation. The interface allows annotators to directly paint impacted regions on the mesh surface and export consistent vertex-level labels across frames. Additional details on the interface design and annotation workflow are provided in Section~\ref{sec:spatial_annotation_interface}.

\paragraph{Annotation Quality Assessment.}
We evaluate annotation quality through inter-annotator agreement and reference-based validation, following~\citet{tripathi2023deco}. These evaluations examine annotator consistency in temporal boundaries and spatial labels, including challenging cases with occlusions and diverse body poses. Detailed procedures and results are provided in Section~\ref{sec:quality_control_assessment}.

\begin{table*}[t]
\centering
\scriptsize
\setlength{\tabcolsep}{3.2pt}
\caption{\textbf{Comparison with Representative Body-Centric Physical Understanding and Video Moment Retrieval Datasets.} \#Text Queries, Avg. \#M / Query, and Avg. M Length denote the number of queries, average moments per query, and average moment duration, respectively. HAPACT contains shorter moments than existing video grounding datasets, with an average duration of $0.20$s.}
\label{tab:dataset_comparison}
\resizebox{\textwidth}{!}{
\begin{tabular}{llcccccc}
\toprule[1.5pt]
\textbf{Dataset} & \textbf{Domain} & \textbf{Scale} & \shortstack{\textbf{\#Text}\\\textbf{Queries}} & \shortstack{\textbf{Avg. \#M}\\\textbf{/ Query}} & \shortstack{\textbf{Avg. M}\\\textbf{Length}} & \textbf{Physical Cue} & \textbf{Body Rep.} \\
 \midrule
\multicolumn{8}{l}{\color{gray}\textit{Image/video body-level physical datasets}} \\
PROX~\citep{hassan2019resolving} & Indoor scenes & 100,000 frames & \xmark & \xmark & \xmark & Scene contact & SMPL-X \\
BEHAVE~\citep{bhatnagar2022behave} & Natural scenes & 15,200 frames & \xmark & \xmark & \xmark & Object contact & SMPL-H \\
RICH~\citep{huang2022capturing} & Natural scenes & 577,000 frames & \xmark & \xmark & \xmark & Scene contact & SMPL-X \\
HOT~\citep{chen2023detecting} & In-the-wild & 15,082 images & \xmark & \xmark & \xmark & Object/scene contact & 2D regions \\
DAMON~\citep{tripathi2023deco} & In-the-wild & 5,522 images & \xmark & \xmark & \xmark & Object/scene contact & SMPL \\
3DIR~\citep{yang2024lemon} & In-the-wild & 5,000 images & \xmark & \xmark & \xmark & Object contact & SMPL-H \\
BodyContact4D~\citep{shinbodycontact4d} & In-the-wild & 800,000 frames & \xmark & \xmark & \xmark & Object/scene contact & 24 segments \\
\midrule
\multicolumn{8}{l}{\color{gray}\textit{Video moment retrieval datasets}} \\
TACoS~\citep{regneri2013grounding} & Cooking & 127 videos & 18,818 & 1 & 6.1s & \xmark & \xmark \\
Charades-STA~\citep{gao2017tall} & Indoor activities & 9,848 videos & 16,128 & 1 & 8.1s & \xmark & \xmark \\
DiDeMo~\citep{anne2017localizing} & Open videos & 10,464 videos & 40,543 & 1 & 6.5s & \xmark & \xmark \\
ActivityNet Captions~\citep{krishna2017dense} & Open videos & 19,811 videos & 71,957 & 1 & 37.1s & \xmark & \xmark \\
TVR~\citep{lei2020tvr} & TV shows & 21,800 videos & 108,965 & 1 & 9.1s & \xmark & \xmark \\
QVHighlights~\citep{lei2021detecting} & Vlog / news & 10,148 videos & 10,310 & 1.78 & 24.6s & \xmark & \xmark \\
MAD~\citep{soldan2022mad} & Movies & 650 movies & 384,600 & 1 & 4.1s & \xmark & \xmark \\
Ego4D-NLQ~\citep{grauman2022ego4d} & Egocentric & 1,659 videos & 19,200 & 1 & 10.5s & \xmark & \xmark \\
\midrule
\textbf{HAPACT} & \textbf{Movies} & \shortstack{\textbf{20,369 annotated frames}\\\textbf{from 2,884 videos}} & \textbf{2,884} & \textbf{1.47} & \shortstack{\textbf{0.2s}\\\textbf{(4.80 f)}} & \shortstack{\textbf{Text-conditioned}\\\textbf{impact events}} & \textbf{SMPL-X} \\
\bottomrule[1.5pt]
\end{tabular}
}
\end{table*}

\begin{figure*}[t]
  \centering
  \includegraphics[width=\textwidth]{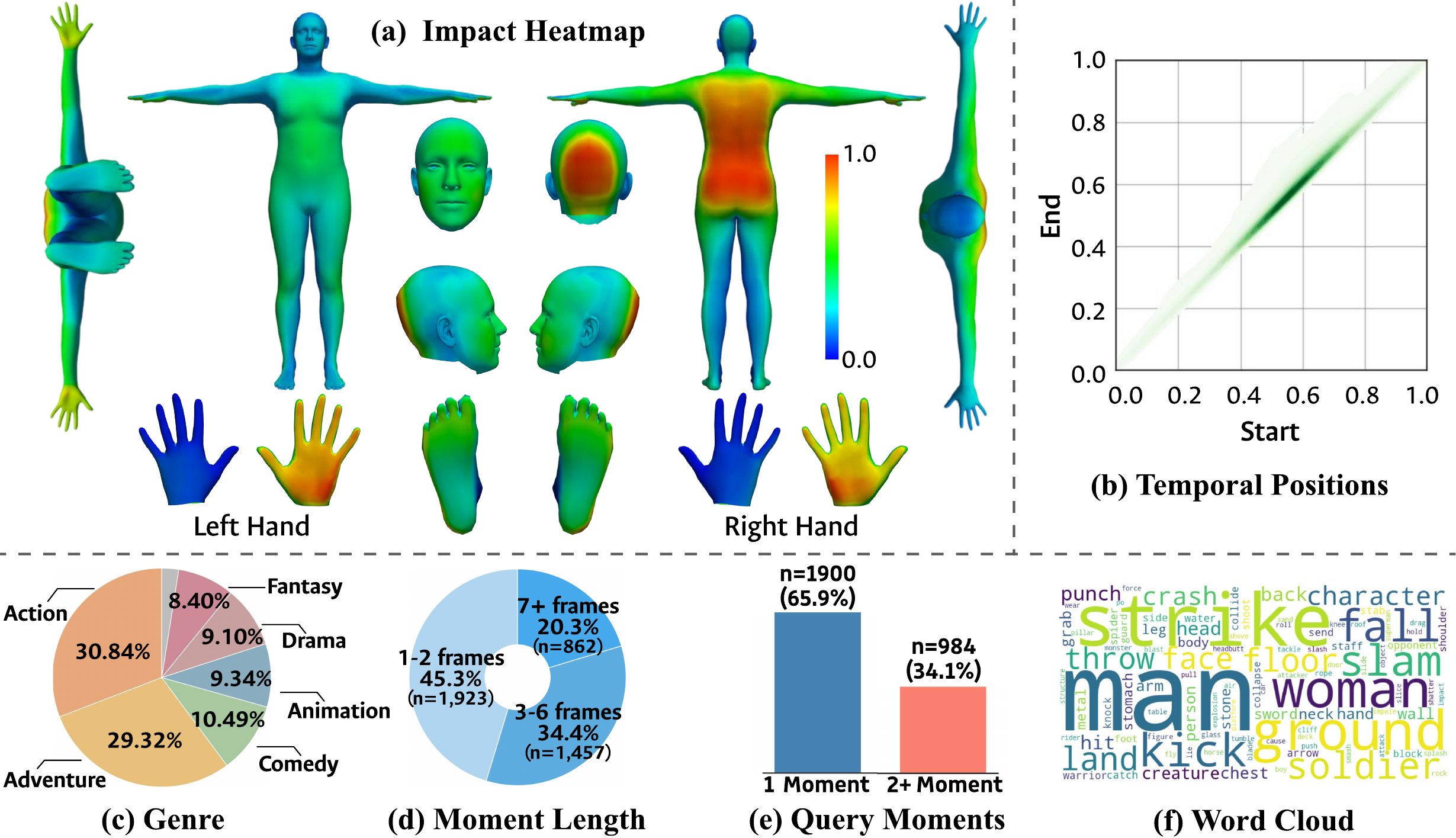}
  \caption{\textbf{Data Distribution Overview.} 
  (a) Distribution of impact occurrences on body mesh, where 1 indicates high and 0 indicates low occurrence probability;
  (b) Distribution of start-end positions of impact moments, normalized from 0 to 1 relative to the total video length;
  (c) Distribution of movie genres;
  (d) Distribution of moment lengths by frame count;   
  (e) Distribution of single- and multi-moment queries; 
  (f) Word cloud of verbs and nouns in the description of HAPACT. It shows that verbs such as ``strike'' and nouns such as ``man'' appear frequently, emphasizing that the dataset focuses on physical interactions. More detailed statistical information can be found in Section~\ref{sec:statistics_details}.}
  \label{fig:analysis}
\end{figure*}

\subsection{Dataset Analysis}

\paragraph{Statistics.}
To the best of our knowledge, HAPACT is the first human-centric video dataset to provide fine-grained spatio-temporal annotations for physical impact events. Table~\ref{tab:dataset_comparison} compares its statistics with existing datasets. Unlike prior datasets, it combines text-conditioned temporal grounding with frame-level on-body impact annotations.

\paragraph{Distribution of Movie Clips and Queries.}
In total, HAPACT contains $2,884$ video-query pairs with $4,242$ moments, drawn from $212$ movies. The dataset is designed around short and salient physical impacts, which is reflected in its temporal statistics. The average moment length is $4.80$ frames, corresponding to approximately $0.20$ seconds at $24$~FPS, and most moments span $1$ to $6$ frames as shown in Fig.~\ref{fig:analysis}(d). A single video-query pair can contain repeated occurrences of the queried impact, resulting in an average of $1.47$ moments per query; as shown in Fig.~\ref{fig:analysis}(e), multi-moment cases account for $34.1\%$ of the dataset. The relative start and end positions of these impact moments are shown in Fig.~\ref{fig:analysis}(b).

The source movie distribution is shaped by the impact-centered construction of HAPACT. As shown in Fig.~\ref{fig:analysis}(c), action and adventure genres appear frequently in the collection. Fig.~\ref{fig:analysis}(f) shows a word cloud of verbs and nouns appearing in the query sentences, which contain $11.1$ words on average. 

\paragraph{Distribution of Impact Occurrence on Body Mesh.}
Fig.~\ref{fig:analysis}(a) shows the aggregate vertex-level impact distribution on the SMPL-X mesh across the entire dataset. The head region has the highest intensity (23.3\%), followed by the hips (8.4\%) and the spine (6.9\%). The body-part visualizations in the same panel further show that impacts frequently occur on the back, buttocks, and palms. These patterns reflect common physical impact scenarios in movie clips, such as falls, collisions, and body-ground or body-object contacts.

\subsection{Dataset Split}
\label{sec:dataset_split}

For benchmark evaluation, we partition HAPACT into training, validation, and test splits with a ratio of $70$:$15$:$15$. The split is movie-disjoint to prevent content leakage, and is constructed to maintain comparable distributions of genre, clip duration, number of moments, moment length, and affected body vertices. The resulting splits contain $2,019$ training pairs with $14,264$ annotated frames, $432$ validation pairs with $2,963$ annotated frames, and $433$ test pairs with $3,142$ annotated frames. Section~\ref{sec:statistics_overall} reports detailed statistics for each split.

\section{Experiments}

\subsection{Task Definition}
\label{sec:task_definition}

Human-centric impact localization is defined over a video with frames $V=\{f_t\}_{t=1}^{T}$ and a textual query $q$ describing a physical impact event involving a human. The objective is to localize both \emph{when} the queried impact occurs and \emph{which body vertices} are affected over time. The task is evaluated along two axes: temporal localization and spatial on-body localization.

\paragraph{Task 1: Temporal Localization.}
The first task is to predict a set of temporal intervals:
\[
\mathcal{T}=\{\tau_k\}_{k=1}^{K}, \quad \tau_k=[t_s^{(k)}, t_e^{(k)}], \quad 1 \le t_s^{(k)} \le t_e^{(k)} \le T,
\]
where each interval corresponds to one occurrence of the queried impact event in the video.

\paragraph{Task 2: Spatial On-body Localization.}
The second task is to predict, for each frame $t \in \bigcup_{k=1}^{K}\tau_k$, a binary on-body label:
\[
\mathbf{y}_t \in \{0,1\}^{N},
\]
where $N$ denotes the number of vertices in the SMPL-X body mesh and $\mathbf{y}_t^{(i)}=1$ indicates that the $i$-th vertex is affected by the queried impact at frame $t$.

\subsection{Benchmark Setup}
\label{sec:benchmark_setup}

HAPACT provides two benchmark axes based on the task definition: \textbf{temporal impact localization} and \textbf{spatial on-body impact localization}. In the following sections, we describe the specialized baselines and evaluation metrics for each task, with implementation details included in Section~\ref{sec:evaluation_details}.

\subsubsection{Temporal Evaluation}

\paragraph{Models.}
 The temporal benchmark includes two types of video moment retrieval~(VMR) models: retrieval-based temporal localization models and generation-based models. Retrieval-based models~\citep{huang2024semantic, lei2021detecting, liu2024r} predict temporal intervals by matching video representations with the query, whereas generation-based models~\citep{huang2024vtimellm, liuniversal, meinardus2025chrono} directly produce temporal predictions from the video using prompts that incorporate the query text. Overall, we evaluate $18$ temporal baselines, comprising 8 retrieval-based models~\citep{huang2024semantic, jang2023knowing, lei2021detecting, moon2023correlation, moon2023query, sun2024tr, xiao2024bridging, tang2025sim} capable of multi-moment prediction and $10$ generation-based models spanning VMR-specific models~\citep{guo2024trace, huang2024vtimellm, liuniversal, wang-etal-2025-grounded, zhang2025timelens}, open-source MLLMs~\citep{bai2025qwen3, wang2025internvl3}, and closed-source MLLMs~\citep{comanici2025gemini, hurst2024gpt, openai2026gpt54systemcard}.

\paragraph{Metrics.}
We mainly evaluate temporal localization performance using conventional VMR metrics~\citep{lei2021detecting}, mean average precision~(mAP) and Recall@1~(R1), at a temporal intersection-over-union~(tIoU) threshold of $0.3$. We use this threshold because impact moments in HAPACT are short, lasting only $0.20$ seconds on average, and are sensitive to small boundary shifts. To account for multi-moment queries, we also report the standard metrics adopted in \citet{caoone}: generalized mean average precision~(G-mAP) and mean recall~(mR), computed over tIoU thresholds from $0.1$ to $0.5$ with a step size of $0.1$. G-mAP metrics are further reported for single-moment~(@1\_tgt), two-moment~(@2\_tgt), and three-or-more-moment~(@3+tgt) queries.

\begin{wrapfigure}{r}{0.575\textwidth} 
    \vspace{-20pt}
    \centering
    \includegraphics[width=\linewidth]{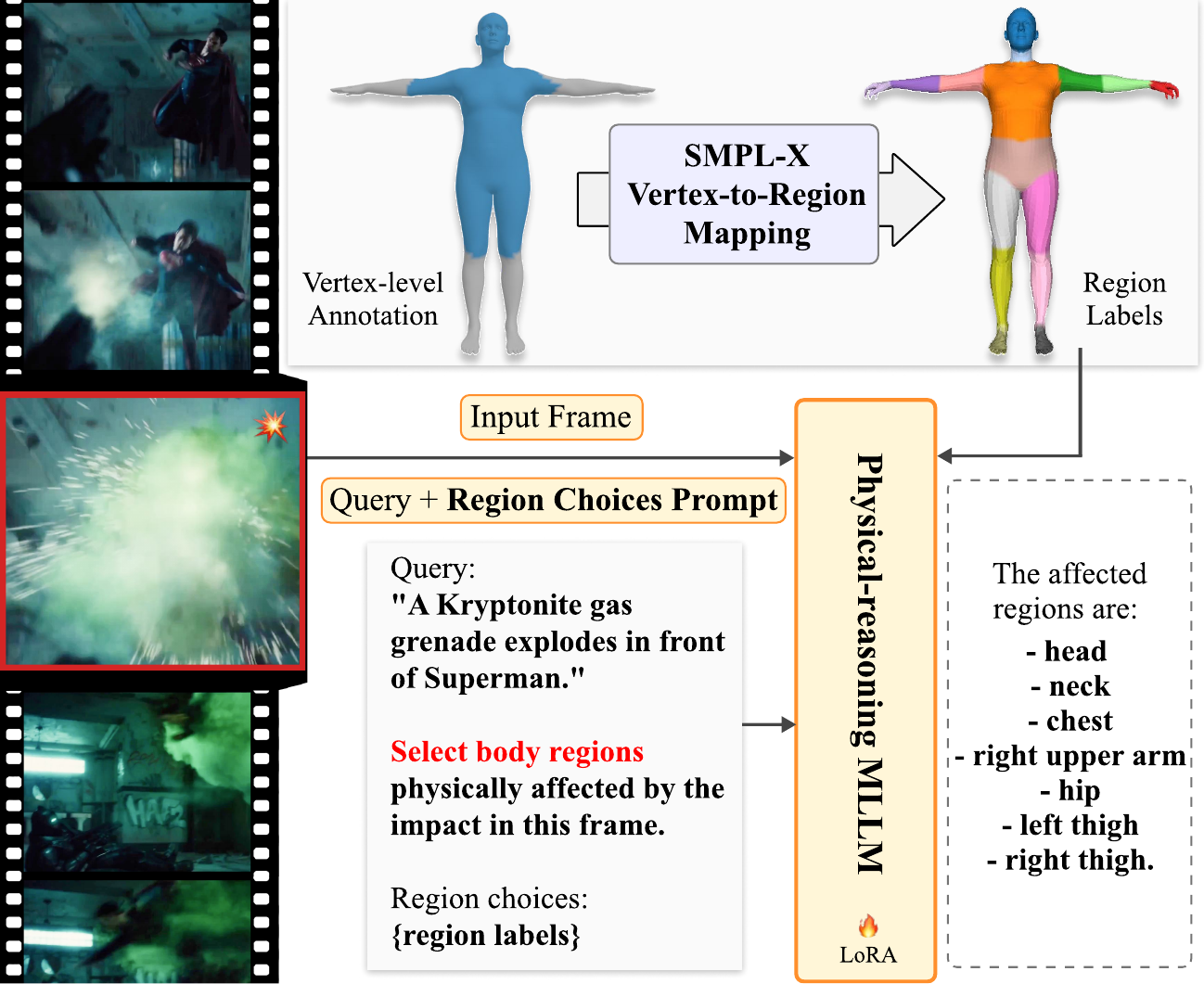}
    \caption{MLLM Baseline for Spatial Impact Localization with Query-Guided Region Selection.}
    \vspace{-25pt}
\end{wrapfigure}

\subsubsection{Spatial Evaluation}

\paragraph{Models.}
The spatial benchmark includes three human-scene contact (HSC) baselines: BSTRO~\citep{huang2022capturing}, DECO~\citep{tripathi2023deco}, and GraphiContact~\citep{lin2026graphicontact}. These models estimate body regions in contact from visual input, and are adopted here to examine whether contact representations can transfer to impact localization.

\paragraph{MLLM Baseline.}
Since the spatial axis lacks established baselines that incorporate text input, we introduce a text-conditioned MLLM baseline that formulates spatial impact localization as query-guided region selection. Inspired by prior studies~\citep{chen2023detecting, shinbodycontact4d}, we convert dense binary labels on SMPL-X into 18 body region groups and provide them as candidate region choices in the prompt. The model is asked to select all regions physically affected by the given impact query. We instantiate this formulation with Cosmos-Reason2~\citep{nvidia2025cosmosreason2}, an MLLM specialized for physical reasoning, and evaluate it after instruction fine-tuning on our training set.

\begin{table}[t]
\centering
\caption{\textbf{Performance Comparison on Temporal Impact Localization of Retrieval-based Moment Retrieval Models.} \emph{ZS} denotes zero-shot evaluation using QVHighlights~\citep{lei2021detecting} checkpoints, and \emph{FT} denotes fine-tuning on the HAPACT training split. \textcolor{gray}{\emph{Random}} evaluates randomly sampled intervals within each clip. \textbf{Bold} indicates the best result and \underline{underline} indicates the second-best result.} 
\label{tab:direct_vmr_models}
\resizebox{\textwidth}{!}{
\begin{tabular}{@{}l crr rrrr rrr@{}}
\toprule[1.5pt]
\multirow{2}{*}{Method} & \multicolumn{3}{c}{} & \multicolumn{4}{c}{mAP} & \multicolumn{3}{c}{mR@k} \\
\cmidrule(lr){5-8} \cmidrule(lr){9-11}
& Setting & mAP@0.3 & R1@0.3 & G-mAP & @1\_tgt & @2\_tgt & @3+tgt & @1 & @2 & @3 \\
 \midrule
\textcolor{gray}{\emph{Random}} & & \textcolor{gray}{4.14} & \textcolor{gray}{2.31} & \textcolor{gray}{7.16} & \textcolor{gray}{7.30} & \textcolor{gray}{6.73} & \textcolor{gray}{5.50} & \textcolor{gray}{4.11} & \textcolor{gray}{6.55} & \textcolor{gray}{7.91} \\
\midrule
\multirow{2}{*}{Moment-DETR~\citep{lei2021detecting}~\scriptsize{\textit{NeurIPS '21}}} & \emph{ZS} & 8.05 & 5.77 & 10.04 & 11.69 & 7.07 & 5.56 & 6.93 & 5.88 & 7.29 \\
& \cellcolor[HTML]{E0E0E0}\emph{FT} & \cellcolor[HTML]{E0E0E0}18.16 & \cellcolor[HTML]{E0E0E0}13.63 & \cellcolor[HTML]{E0E0E0}20.39 & \cellcolor[HTML]{E0E0E0}21.94 & \cellcolor[HTML]{E0E0E0}19.11 & \cellcolor[HTML]{E0E0E0}12.92 & \cellcolor[HTML]{E0E0E0}15.57 & \cellcolor[HTML]{E0E0E0}16.61 & \cellcolor[HTML]{E0E0E0}16.59 \\
\midrule
\multirow{2}{*}{QD-DETR~\citep{moon2023query}~\scriptsize{\textit{CVPR '23}}} & \emph{ZS} & 9.55 & 6.00 & 11.98 & 13.57 & 9.50 & 7.81 & 8.22 & 8.12 & 8.53 \\
& \cellcolor[HTML]{E0E0E0}\emph{FT} & \cellcolor[HTML]{E0E0E0}21.27 & \cellcolor[HTML]{E0E0E0}\underline{16.17} & \cellcolor[HTML]{E0E0E0}22.06 & \cellcolor[HTML]{E0E0E0}23.91 & \cellcolor[HTML]{E0E0E0}19.98 & \cellcolor[HTML]{E0E0E0}15.39 & \cellcolor[HTML]{E0E0E0}16.86 & \cellcolor[HTML]{E0E0E0}18.67 & \cellcolor[HTML]{E0E0E0}19.69 \\
\midrule
\multirow{2}{*}{EaTR~\citep{jang2023knowing}~\scriptsize{\textit{ICCV '23}}} & \emph{ZS} & 9.49 & 6.24 & 12.28 & 12.85 & 11.78 & 6.29 & 8.13 & 9.64 & 11.01 \\
& \cellcolor[HTML]{E0E0E0}\emph{FT} & \cellcolor[HTML]{E0E0E0}\underline{21.69} & \cellcolor[HTML]{E0E0E0}\underline{16.17} & \cellcolor[HTML]{E0E0E0}22.54 & \cellcolor[HTML]{E0E0E0}23.09 & \cellcolor[HTML]{E0E0E0}22.58 & \cellcolor[HTML]{E0E0E0}13.92 & \cellcolor[HTML]{E0E0E0}16.95 & \cellcolor[HTML]{E0E0E0}19.21 & \cellcolor[HTML]{E0E0E0}20.62 \\
\midrule
\multirow{2}{*}{CG-DETR~\citep{moon2023correlation}~\scriptsize{\textit{arXiv '24}}}& \emph{ZS} & 7.96 & 4.62 & 11.09 & 13.05 & 8.27 & 5.08 & 7.11 & 6.06 & 7.29 \\
& \cellcolor[HTML]{E0E0E0}\emph{FT} & \cellcolor[HTML]{E0E0E0}20.57 & \cellcolor[HTML]{E0E0E0}15.47 & \cellcolor[HTML]{E0E0E0}22.55 & \cellcolor[HTML]{E0E0E0}23.96 & \cellcolor[HTML]{E0E0E0}19.48 & \cellcolor[HTML]{E0E0E0}\underline{17.28} & \cellcolor[HTML]{E0E0E0}17.92 & \cellcolor[HTML]{E0E0E0}\underline{21.52} & \cellcolor[HTML]{E0E0E0}22.63 \\
\midrule
\multirow{2}{*}{SFABD~\citep{huang2024semantic}~\scriptsize{\textit{WACV '24}}} & \emph{ZS} & 5.14 & 2.08 & 7.59 & 8.52 & 5.82 & 4.12 & 3.56 & 4.55 & 6.20 \\
& \cellcolor[HTML]{E0E0E0}\emph{FT} & \cellcolor[HTML]{E0E0E0}21.19 & \cellcolor[HTML]{E0E0E0}15.70 & \cellcolor[HTML]{E0E0E0}20.59 & \cellcolor[HTML]{E0E0E0}21.91 & \cellcolor[HTML]{E0E0E0}19.54 & \cellcolor[HTML]{E0E0E0}12.59 & \cellcolor[HTML]{E0E0E0}15.79 & \cellcolor[HTML]{E0E0E0}15.70 & \cellcolor[HTML]{E0E0E0}17.21 \\
\midrule
\multirow{2}{*}{TR-DETR~\citep{sun2024tr}~\scriptsize{\textit{AAAI '24}}} & \emph{ZS} & 10.62 & 7.39 & 14.00 & 14.97 & 12.81 & 6.99 & 10.21 & 9.64 & 11.01 \\
& \cellcolor[HTML]{E0E0E0}\emph{FT} & \cellcolor[HTML]{E0E0E0}20.86 & \cellcolor[HTML]{E0E0E0}\underline{16.17} & \cellcolor[HTML]{E0E0E0}22.20 & \cellcolor[HTML]{E0E0E0}23.86 & \cellcolor[HTML]{E0E0E0}20.36 & \cellcolor[HTML]{E0E0E0}14.62 & \cellcolor[HTML]{E0E0E0}17.69 & \cellcolor[HTML]{E0E0E0}17.94 & \cellcolor[HTML]{E0E0E0}21.39 \\
\midrule
\multirow{2}{*}{UVCOM~\citep{xiao2024bridging}~\scriptsize{\textit{CVPR '24}}} & \emph{ZS} & 10.20 & 7.39 & 13.65 & 15.56 & 10.71 & 5.98 & 10.67 & 8.12 & 8.22 \\
& \cellcolor[HTML]{E0E0E0}\emph{FT} & \cellcolor[HTML]{E0E0E0}21.22 & \cellcolor[HTML]{E0E0E0}\textbf{17.32} & \cellcolor[HTML]{E0E0E0}\underline{23.58} & \cellcolor[HTML]{E0E0E0}\underline{24.28} & \cellcolor[HTML]{E0E0E0}\underline{22.96} & \cellcolor[HTML]{E0E0E0}16.53 & \cellcolor[HTML]{E0E0E0}\textbf{19.26} & \cellcolor[HTML]{E0E0E0}\textbf{21.76} & \cellcolor[HTML]{E0E0E0}\underline{24.65} \\
\midrule
\multirow{2}{*}{Sim-DETR~\citep{tang2025sim}~\scriptsize{\textit{ICCV '25}}} & \emph{ZS} & 5.67 & 1.85 & 7.31 & 7.66 & 6.75 & 6.49 & 3.23 & 4.79 & 6.36 \\
& \cellcolor[HTML]{E0E0E0}\emph{FT} & \cellcolor[HTML]{E0E0E0}\textbf{23.27} & \cellcolor[HTML]{E0E0E0}15.47 & \cellcolor[HTML]{E0E0E0}\textbf{26.07} & \cellcolor[HTML]{E0E0E0}\textbf{26.58} & \cellcolor[HTML]{E0E0E0}\textbf{24.23} & \cellcolor[HTML]{E0E0E0}\textbf{25.41} & \cellcolor[HTML]{E0E0E0}\underline{18.62} & \cellcolor[HTML]{E0E0E0}\textbf{21.76} & \cellcolor[HTML]{E0E0E0}\textbf{26.82} \\
\bottomrule[1.5pt]
\end{tabular}
}
\end{table}

\begin{table}[t]
\centering
\caption{\textbf{Zero-shot Performance Comparison on Temporal Impact Localization of Generation-based Moment Retrieval Models.} 
Closed-source MLLM results are reported as mean (standard deviation) over five runs.  Detailed evaluation settings are provided in Section~\ref{sec:temporal_evaluation_details}.
{\setlength{\fboxsep}{1pt}\colorbox[HTML]{FCD6B4}{\textbf{Orange}}} denotes the best result, while 
{\setlength{\fboxsep}{1pt}\colorbox[HTML]{CBCEFB}{purple}} indicates the second best.}
\label{tab:gen_vmr_models}
\resizebox{\textwidth}{!}{
% \begin{tabular}{@{}l c | rr rrrr rrr@{}}
\begin{tabular}{@{}l rr rrrr rrr@{}}
\toprule[1.5pt]
\multirow{2}{*}{Method} & \multicolumn{2}{c}{} & \multicolumn{4}{c}{mAP} & \multicolumn{3}{c}{mR@k} \\
\cmidrule(lr){4-7} \cmidrule(lr){8-10}
 & mAP@0.3 & R1@0.3 & G-mAP & @1\_tgt & @2\_tgt & @3+tgt & @1 & @2 & @3 \\
 \midrule
\multicolumn{10}{@{}l}{\color{gray}~\scriptsize{\textit{VMR-specific Models}}} \\
VTimeLLM~\citep{huang2024vtimellm} & 1.04 & 1.15 & 1.92 & 2.69 & 0.82 & 0.26 & 2.22 & 0.00 & 0.00 \\
TRACE~\citep{guo2024trace} & 4.21 & 4.16 & 4.76 & 5.18 & 3.67 & 3.25 & 4.80 & 6.25 & 7.50 \\
Grounded-Video-LLM~\citep{wang-etal-2025-grounded} & 7.97 & 10.62 & 10.06 & 11.72 & 8.44 & 1.37 & 13.30 & 0.00 & 0.00 \\
UniTime~\citep{liuniversal} & 7.93 & 9.01 & 12.41 & 16.42 & 6.72 & 1.95 & 15.06 & 0.00 & 0.00 \\
TimeLens~\citep{zhang2025timelens} & 5.10 & 5.77 & 9.60 & 12.54 & 5.08 & 2.95 & 11.83 & 5.00 & 0.00 \\
\midrule
\multicolumn{10}{@{}l}{\color{gray}~\scriptsize{\textit{Open Source General MLLMs}}} \\
Qwen3-VL-8B~\citep{bai2025qwen3} & 22.88 & 29.33 & 23.25 & 28.31 & 16.23 & 7.08 & 30.07 & 7.69 & 11.11 \\
Qwen3-VL-32B~\citep{bai2025qwen3} & 21.45 & 24.48 & 22.62 & 28.90 & 13.66 & 5.63 & 26.50 & 19.14 & 6.67 \\
InternVL3.5-8B~\citep{wang2025internvl3} & 14.01 & 16.17 & 13.67 & 14.12 & 14.63 & 5.27 & 15.98 & 21.18 & 15.00 \\
% InternVL3.5-38B~\scriptsize{\textit{arXiv '25}} & 192 & -- & -- & -- & -- & -- & -- & -- & -- & -- \\
\midrule
\multicolumn{10}{@{}l}{\color{gray}~\scriptsize{\textit{Closed Source General MLLMs}}} \\
Gemini 3.1 Pro~\citep{google2026gemini31pro} & 27.22 {\scriptsize(1.34)} & 29.93 {\scriptsize(1.47)} & 29.70 {\scriptsize(1.05)} & \cellcolor[HTML]{CBCEFB}38.38 {\scriptsize(1.89)} & 16.87 {\scriptsize(1.33)} & 7.21 {\scriptsize(1.94)} & 33.68 {\scriptsize(1.28)} & \cellcolor[HTML]{CBCEFB}28.11 {\scriptsize(1.96)} & \cellcolor[HTML]{CBCEFB}22.56 {\scriptsize(4.65)} \\
GPT-4o~\citep{hurst2024gpt} & \cellcolor[HTML]{CBCEFB}28.95 {\scriptsize(1.22)} & \cellcolor[HTML]{CBCEFB}33.67 {\scriptsize(1.63)} & \cellcolor[HTML]{CBCEFB}30.10 {\scriptsize(0.76)} & 37.97 {\scriptsize(1.20)} & \cellcolor[HTML]{CBCEFB}18.30 {\scriptsize(0.97)} & \cellcolor[HTML]{CBCEFB}8.58 {\scriptsize(1.64)} & \cellcolor[HTML]{CBCEFB}34.97 {\scriptsize(0.85)} & 22.62 {\scriptsize(2.53)} & 17.69 {\scriptsize(9.63)} \\
GPT-5.4~\citep{openai2026gpt54systemcard} & \cellcolor[HTML]{FCD6B4}\textbf{42.64} {\scriptsize(0.86)} & \cellcolor[HTML]{FCD6B4}\textbf{47.76} {\scriptsize(0.69)} & \cellcolor[HTML]{FCD6B4}\textbf{43.64} {\scriptsize(0.74)} & \cellcolor[HTML]{FCD6B4}\textbf{54.49} {\scriptsize(1.30)} & \cellcolor[HTML]{FCD6B4}\textbf{26.90} {\scriptsize(1.39)} & \cellcolor[HTML]{FCD6B4}\textbf{14.74} {\scriptsize(1.88)} & \cellcolor[HTML]{FCD6B4}\textbf{48.80} {\scriptsize(0.62)} & \cellcolor[HTML]{FCD6B4}\textbf{38.89} {\scriptsize(1.20)} & \cellcolor[HTML]{FCD6B4}\textbf{33.92} {\scriptsize(5.02)} \\
\bottomrule[1.5pt]
\end{tabular}
}
\end{table}

\paragraph{Metrics.}
Following prior human-scene contact estimation works~\citep{huang2022capturing, lin2026graphicontact, tripathi2023deco}, spatial localization performance is evaluated using frame-wise precision, recall, and F1 score, together with geodesic error in centimeters. For analysis by region, we construct seven body region groups by merging the part segmentation from \citet{chen2023detecting}: head, torso, arms, legs, hips, hands, and feet. For MLLM evaluation, since predictions are made over candidate region labels, we report F1 scores over the predefined labels. Further details are provided in Section~\ref{sec:spatial_localization_evaluation}.

\subsection{Main Results}
\label{sec:main_results}

\paragraph{Fine-Tuning on HAPACT improves retrieval-based models on temporal impact localization.}
For retrieval-based models, Table~\ref{tab:direct_vmr_models} shows that QVHighlights pre-training does not consistently carry over to temporal impact localization, with several models remaining close to the random baseline. Fine-tuning on HAPACT improves all retrieval models, suggesting that existing VMR architectures can learn to localize impact events with temporal supervision. Still, the best scores remain moderate, reflecting the difficulty of precise localization under short durations and multiple annotated moments.

\paragraph{General MLLMs localize salient impacts but struggle with multi-moments.}
As presented in Table~\ref{tab:gen_vmr_models}, VMR-targeted generation-based models remain weak on temporal impact localization. In contrast, general MLLMs achieve much stronger zero-shot performance, showing that broad video-language understanding helps identify human impact events. However, even the strongest model (GPT-5.4) drops from mAP@1\_tgt to mAP@2\_tgt and mAP@3+tgt, decreasing by about $51\%$ and $73\%$, respectively. This decline indicates that general MLLMs can identify salient impact occurrences, but remain limited in recovering multiple target moments with accurate temporal boundaries.

\begin{table}[t]
\centering
\caption{\textbf{Performance Comparison on Spatial On-body Impact Localization.}  \emph{ZS}, \emph{FT}, and \emph{RS} denote zero-shot, fine-tuning on HAPACT, and fine-tuning with query-guided region selection, respectively. Per-region F1 is reported over seven region groups. \textbf{Bold} indicates the best result.}
\label{tab:human_scene_contact}
\resizebox{\textwidth}{!}{
\begin{tabular}{@{}l crrrr rrrrrrr@{}}
\toprule[1.5pt]
\multirow{2}{*}{Method} & \multirow{2}{*}{Setting} & \multirow{2}{*}{F1 $\uparrow$} & \multirow{2}{*}{Precision $\uparrow$} & \multirow{2}{*}{Recall $\uparrow$} & \multirow{2}{*}{geo. (cm) $\downarrow$} & \multicolumn{7}{c}{F1 Scores per Body-part $\uparrow$} \\
\cmidrule(lr){7-13}
 & & & & & & Head & Arms & Torso & Hips & Legs & Hands & Feet \\
 \midrule
\multirow{2}{*}{BSTRO~\citep{huang2022capturing}} & \emph{ZS} & 6.13 & 10.69 & 5.68 & 81.53 & 0.00 & 2.75 & 0.00 & 0.35 & 0.01 & 14.82 & 16.61 \\
 & \cellcolor[HTML]{F0F0F0}\emph{FT} & \cellcolor[HTML]{F0F0F0}5.72 & \cellcolor[HTML]{F0F0F0}11.20 & \cellcolor[HTML]{F0F0F0}5.93 & \cellcolor[HTML]{F0F0F0}51.35 & \cellcolor[HTML]{F0F0F0}1.65 & \cellcolor[HTML]{F0F0F0}3.72 & \cellcolor[HTML]{F0F0F0}7.44 & \cellcolor[HTML]{F0F0F0}13.40 & \cellcolor[HTML]{F0F0F0}4.30 & \cellcolor[HTML]{F0F0F0}7.18 & \cellcolor[HTML]{F0F0F0}3.66 \\
\midrule
\multirow{2}{*}{DECO~\citep{tripathi2023deco}} & \emph{ZS} & 8.28 & 9.87 & 11.84 & 83.84 & 0.00 & 0.25 & 1.45 & 4.21 & 1.26 & 12.87 & 28.86 \\
 & \cellcolor[HTML]{F0F0F0}\emph{FT} & \cellcolor[HTML]{F0F0F0}5.41 & \cellcolor[HTML]{F0F0F0}\textbf{68.16} & \cellcolor[HTML]{F0F0F0}6.18 & \cellcolor[HTML]{F0F0F0}50.34 & \cellcolor[HTML]{F0F0F0}1.46 & \cellcolor[HTML]{F0F0F0}4.78 & \cellcolor[HTML]{F0F0F0}8.79 & \cellcolor[HTML]{F0F0F0}10.80 & \cellcolor[HTML]{F0F0F0}0.85 & \cellcolor[HTML]{F0F0F0}6.77 & \cellcolor[HTML]{F0F0F0}0.20 \\
\midrule
\multirow{2}{*}{GraphiContact~\citep{lin2026graphicontact}} & \emph{ZS} & 7.88 & 8.85 & 10.23 & 87.42 & 0.09 & 0.28 & 1.34 & 8.52 & 1.84 & 0.63 & 29.34 \\
 & \cellcolor[HTML]{F0F0F0}\emph{FT} & \cellcolor[HTML]{F0F0F0}7.34 & \cellcolor[HTML]{F0F0F0}30.12 & \cellcolor[HTML]{F0F0F0}14.18 & \cellcolor[HTML]{F0F0F0}50.02 & \cellcolor[HTML]{F0F0F0}5.57 & \cellcolor[HTML]{F0F0F0}8.48 & \cellcolor[HTML]{F0F0F0}11.68 & \cellcolor[HTML]{F0F0F0}15.63 & \cellcolor[HTML]{F0F0F0}7.19 & \cellcolor[HTML]{F0F0F0}15.62 & \cellcolor[HTML]{F0F0F0}6.40 \\
\midrule[1.2pt]
Cosmos-Reason2-2B \multirow{2}{*}{\citep{nvidia2025cosmosreason2}}& \cellcolor[HTML]{E0E0E0}\emph{RS} & \cellcolor[HTML]{E0E0E0}13.16 & \cellcolor[HTML]{E0E0E0}11.11 & \cellcolor[HTML]{E0E0E0}32.41 & \cellcolor[HTML]{E0E0E0}52.34 & \cellcolor[HTML]{E0E0E0}10.99 & \cellcolor[HTML]{E0E0E0}\textbf{16.48} & \cellcolor[HTML]{E0E0E0}16.13 & \cellcolor[HTML]{E0E0E0}18.39 & \cellcolor[HTML]{E0E0E0}10.08 & \cellcolor[HTML]{E0E0E0}\textbf{23.45} & \cellcolor[HTML]{E0E0E0}6.80\\
Cosmos-Reason2-8B & \cellcolor[HTML]{E0E0E0}\emph{RS} & \cellcolor[HTML]{E0E0E0}\textbf{18.59} & \cellcolor[HTML]{E0E0E0}16.17 & \cellcolor[HTML]{E0E0E0}\textbf{44.11} & \cellcolor[HTML]{E0E0E0}\textbf{33.31} & \cellcolor[HTML]{E0E0E0}\textbf{21.21} & \cellcolor[HTML]{E0E0E0}15.92 & \cellcolor[HTML]{E0E0E0}\textbf{20.01} & \cellcolor[HTML]{E0E0E0}\textbf{29.95} & \cellcolor[HTML]{E0E0E0}\textbf{12.59} & \cellcolor[HTML]{E0E0E0}17.87 & \cellcolor[HTML]{E0E0E0}\textbf{8.71} \\
\bottomrule[1.5pt]
\end{tabular}
}
\end{table}

\begin{figure*}[t]
\centering

\begin{minipage}[t]{0.35\textwidth}
\centering
\includegraphics[width=\linewidth]{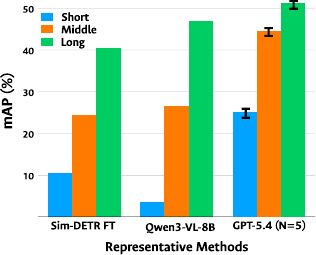}
\captionof{figure}{Temporal mAP scores across impact duration groups for representative models.}
\label{fig:length_wise_map}
\end{minipage}
\hfill
\begin{minipage}[t]{0.60\textwidth}
\centering
\includegraphics[width=\linewidth]{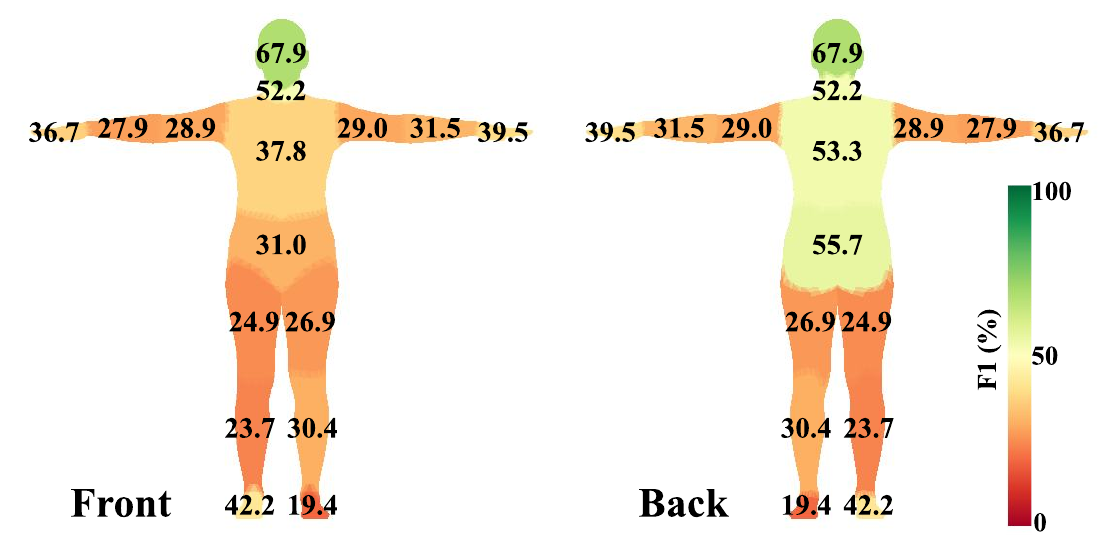}
\captionof{figure}{Label-level F1 scores of Cosmos-Reason2-8B fine-tuned using region selection adaptation. Left and right views show the front and back of the SMPL-X body, respectively.}
\label{fig:cosmos_region_f1}
\end{minipage}

\end{figure*}

\paragraph{Temporal performance degrades on short impact moments.} Figure~\ref{fig:length_wise_map} further compares length-wise mAP for the strongest representative model from each family. Across all models, short moments (1--2 frames) consistently yield lower mAP than middle (3--6 frames) and long moments (7+ frames). This pattern points to a key difficulty in HAPACT, as current moment retrieval models still have difficulty capturing brief physical impacts. Section~\ref{sec:temporal_bottleneck_analysis} presents detailed experiments and analyses of temporal bottlenecks across the evaluated VMR models.

\paragraph{Limited performance of contact-based models on spatial impact localization.} Contact-based baselines achieve low accuracy on spatial on-body impact localization, as reported in Table~\ref{tab:human_scene_contact}. Fine-tuning on HAPACT reduces geodesic errors, but vertex-level F1 scores remain below $9\%$ across all models. These results indicate that physical contact representations alone cannot reliably identify impact-affected body regions from image-only observations. Training from scratch on HAPACT results in similarly low performance, as detailed in Section~\ref{sec:additional_hsc_results}.

\paragraph{Region selection provides a stronger coarse spatial baseline.}
Cosmos-Reason2-8B, fine-tuned with the region selection formulation, obtains the highest spatial scores in Table~\ref{tab:human_scene_contact}, despite using coarse body-region predictions rather than vertex-level labels. Still, Figure~\ref{fig:cosmos_region_f1} shows that its label-level F1 is concentrated on the head, back, and buttocks, while many other regions remain lower. This uneven performance suggests that identifying the affected body region remains challenging even at a coarse level, before considering dense vertex-level localization. Balancing the training-set label distribution narrows but does not eliminate this disparity, as detailed in Section~\ref{sec:region_imbalance_analysis}.

\section{Conclusion}

We introduce human-centric impact localization, a new task for identifying queried physical impacts in time and on the human body. We further present HAPACT, the first human-centric video benchmark with fine-grained annotations for physical impact events. Benchmark results show that current models struggle with brief and multi-moment impacts as well as affected-body prediction. Our region selection formulation enables MLLMs for spatial on-body localization, providing a stronger text-conditioned reference while vertex-level estimation remains challenging. HAPACT extends video-based physical reasoning beyond user-intended interactions to externally driven bodily impacts, with potential applications in body-aware video analysis and media-based haptic generation.

\begin{ack}

The authors declare no competing interests.

\end{ack}

% \newpage
\bibliographystyle{abbrvnat}
\bibliography{ref}

@inproceedings{john2026hapticlens,
  title={HapticLens: Interactive Vibrotactile Haptic Generation from Spatially Localized Video Motion},
  author={John, Kevin and Seifi, Hasti},
  booktitle={Proceedings of the 2026 CHI Conference on Human Factors in Computing Systems},
  pages={1--21},
  year={2026}
}

@article{jung2024hapmotion,
  title={HapMotion: motion-to-tactile framework with wearable haptic devices for immersive VR performance experience},
  author={Jung, Kyungeun and Kim, Sangpil and Oh, Seungjae and Yoon, Sang Ho},
  journal={Virtual Reality},
  volume={28},
  number={1},
  pages={13},
  year={2024},
  publisher={Springer}
}

@inproceedings{li2021towards,
  title={Towards context-aware automatic haptic effect generation for home theatre environments},
  author={Li, Yaxuan and Yoo, Yongjae and Weill-Duflos, Antoine and Cooperstock, Jeremy},
  booktitle={Proceedings of the 27th ACM symposium on virtual reality software and technology},
  pages={1--11},
  year={2021}
}

@article{bansal2026towards,
  title={Towards in-the-wild Egocentric 3D Hand-Object Pose Estimation},
  author={Bansal, Siddhant and Zhu, Zhifan and Tripathi, Shashank and Zhao, Jiahe and Black, Michael J and Damen, Dima},
  journal={arXiv preprint arXiv:2606.30598},
  year={2026}
}

@inproceedings{meinardus2025chrono,
  title={Chrono: A Simple Blueprint for Representing Time in MLLMs},
  author={Meinardus, Boris and Rodriguez, Hector G and Batra, Anil and Rohrbach, Anna and Rohrbach, Marcus},
  booktitle={Proceedings of the IEEE/CVF International Conference on Computer Vision},
  pages={4092--4097},
  year={2025}
}

@inproceedings{liu2024r,
  title={R 2-tuning: Efficient image-to-video transfer learning for video temporal grounding},
  author={Liu, Ye and He, Jixuan and Li, Wanhua and Kim, Junsik and Wei, Donglai and Pfister, Hanspeter and Chen, Chang Wen},
  booktitle={European conference on computer vision},
  pages={421--438},
  year={2024},
  organization={Springer}
}

@inproceedings{sadhu2021visual,
  title={Visual semantic role labeling for video understanding},
  author={Sadhu, Arka and Gupta, Tanmay and Yatskar, Mark and Nevatia, Ram and Kembhavi, Aniruddha},
  booktitle={Proceedings of the IEEE/CVF Conference on Computer Vision and Pattern Recognition},
  pages={5589--5600},
  year={2021}
}

@inproceedings{chen2014fast,
  title={A Fast and Accurate Dependency Parser using Neural Networks},
  author={Chen, Danqi and Manning, Christopher D.},
  booktitle={Proceedings of the 2014 Conference on Empirical Methods in Natural Language Processing (EMNLP)},
  pages={740--750},
  year={2014}
}

@inproceedings{Zala2023HiREST,
  author    = {Abhay Zala and Jaemin Cho and Satwik Kottur and Xilun Chen and Barlas Oğuz and Yashar Mehdad and Mohit Bansal},
  title     = {Hierarchical Video-Moment Retrieval and Step-Captioning},
  booktitle = {CVPR},
  year      = {2023},
}

@ARTICLE{9446308,
  author={Tang, Zongheng and Liao, Yue and Liu, Si and Li, Guanbin and Jin, Xiaojie and Jiang, Hongxu and Yu, Qian and Xu, Dong},
  journal={IEEE Transactions on Circuits and Systems for Video Technology}, 
  title={Human-Centric Spatio-Temporal Video Grounding With Visual Transformers}, 
  year={2022},
  volume={32},
  number={12},
  pages={8238-8249}
}

@inproceedings{zhang2020does,
    title={Where Does It Exist: Spatio-Temporal Video Grounding for Multi-Form Sentences},
    author={Zhang, Zhu and Zhao, Zhou and Zhao, Yang and Wang, Qi and Liu, Huasheng and Gao, Lianli},
    booktitle={CVPR},
    year={2020}
}

@InProceedings{perrett2025hdepic,
  author    = {Perrett, Toby and Darkhalil, Ahmad and Sinha, Saptarshi and Emara, Omar and Pollard, Sam and Parida, Kranti and Liu, Kaiting and Gatti, Prajwal and Bansal, Siddhant and Flanagan, Kevin and Chalk, Jacob and Zhu, Zhifan and Guerrier, Rhodri and Abdelazim, Fahd and Zhu, Bin and Moltisanti, Davide and Wray, Michael and Doughty, Hazel and Damen, Dima},
  title     = {HD-EPIC: A Highly-Detailed Egocentric Video Dataset},
  booktitle = {Proceedings of the IEEE/CVF Conference on Computer Vision and Pattern Recognition (CVPR)},
  year      = {2025},
  month     = {June}
}

@InProceedings{yuan2025momentseeker,
    title={MomentSeeker: A Comprehensive Benchmark and A Strong Baseline For Moment Retrieval Within Long Videos}, 
    author={Huaying Yuan and Jian Ni and Yueze Wang and Junjie Zhou and Zhengyang Liang and Zheng Liu and Zhao Cao and Zhicheng Dou and Ji-Rong Wen},
    year={2025},
    booktitle = {Advances in Neural Information Processing Systems (NeurIPS)},
}

@InProceedings{Croitoru_2023_ICCV,
    author    = {Croitoru, Ioana and Bogolin, Simion-Vlad and Albanie, Samuel and Liu, Yang and Wang, Zhaowen and Yoon, Seunghyun and Dernoncourt, Franck and Jin, Hailin and Bui, Trung},
    title     = {Moment Detection in Long Tutorial Videos},
    booktitle = {Proceedings of the IEEE/CVF International Conference on Computer Vision (ICCV)},
    month     = {October},
    year      = {2023},
    pages     = {2594-2604}
}

@inproceedings{ICLR2025_7ac19fdc,
 author = {You, Qin and Wu, Qilong and Li, Yicong and Ji, Wei and Li, Li and Cai, Pengcheng and Wei, Lina and Zimmermann, Roger},
 booktitle = {International Conference on Learning Representations},
 editor = {Y. Yue and A. Garg and N. Peng and F. Sha and R. Yu},
 pages = {49207--49223},
 title = {Generalized Video Moment Retrieval},
 volume = {2025},
 year = {2025}
}

@InProceedings{cao2025flashmmr,
    author    = {Cao, Zhuo and Du, Heming and Zhang, Bingqing and Yu, Xin and Li, Xue and Wang, Sen},
    title     = {When One Moment Isn't Enough: Multi-Moment Retrieval with Cross-Moment Interactions},
    booktitle = {Advances in Neural Information Processing Systems (NeurIPS)},
    year      = {2025},
}

@article{chen2024verified,
  title={Verified: A video corpus moment retrieval benchmark for fine-grained video understanding},
  author={Chen, Houlun and Wang, Xin and Chen, Hong and Zhang, Zeyang and Feng, Wei and Huang, Bin and Jia, Jia and Zhu, Wenwu},
  journal={Advances in Neural Information Processing Systems},
  volume={37},
  pages={40393--40406},
  year={2024}
}

@inproceedings{yang2024lemon,
  title={Lemon: Learning 3d human-object interaction relation from 2d images},
  author={Yang, Yuhang and Zhai, Wei and Luo, Hongchen and Cao, Yang and Zha, Zheng-Jun},
  booktitle={Proceedings of the IEEE/CVF Conference on Computer Vision and Pattern Recognition},
  pages={16284--16295},
  year={2024}
}

@inproceedings{huang2022intercap,
  title={InterCap: Joint markerless 3D tracking of humans and objects in interaction},
  author={Huang, Yinghao and Taheri, Omid and Black, Michael J and Tzionas, Dimitrios},
  booktitle={DAGM German Conference on Pattern Recognition},
  pages={281--299},
  year={2022},
  organization={Springer}
}

@inproceedings{zhao2025egopressure,
  title={Egopressure: A dataset for hand pressure and pose estimation in egocentric vision},
  author={Zhao, Yiming and Kwon, Taein and Streli, Paul and Pollefeys, Marc and Holz, Christian},
  booktitle={Proceedings of the Computer Vision and Pattern Recognition Conference},
  pages={27727--27738},
  year={2025}
}

@inproceedings{seo2024posture,
  title={Posture-Informed Muscular Force Learning for Robust Hand Pressure Estimation.},
  author={Seo, Kyung Jin and Seo, Junghoon and Jeong, Hanseok and Kim, Sangpil and Yoon, Sang Ho},
  booktitle={NeurIPS},
  year={2024}
}

@inproceedings{grady2022pressurevision,
  title={PressureVision: estimating hand pressure from a single RGB image},
  author={Grady, Patrick and Tang, Chengcheng and Brahmbhatt, Samarth and Twigg, Christopher D and Wan, Chengde and Hays, James and Kemp, Charles C},
  booktitle={European Conference on Computer Vision},
  pages={328--345},
  year={2022},
  organization={Springer}
}

@misc{imdb,
  author       = {{IMDb}},
  title        = {{IMDb} Non-Commercial Datasets},
  year         = {2026},
  howpublished = {\url{https://developer.imdb.com/non-commercial-datasets/}},
  note         = {Accessed: 2026-05-04}
}

@misc{pyscenedetect,
  author       = {Brandon Castellano},
  title        = {{PySceneDetect}: Python and OpenCV-based Scene Cut/Transition Detection Program and Library},
  year         = {2014--2026},
  howpublished = {\url{https://github.com/Breakthrough/PySceneDetect}},
  note         = {Accessed: 2026-05-04}
}

@misc{cj4dx,
  author       = {{CJ 4DPLEX}},
  title        = {{4DX}},
  year         = {n.d.},
  howpublished = {\url{https://www.cj4dplex.com/4dx}},
  note         = {Accessed: 2026-05-04}
}

@misc{labelstudio2025,
  author       = {Tkachenko, Maxim and Malyuk, Mikhail and Holmanyuk, Andrey and Liubimov, Nikolai},
  title        = {{Label Studio}: Data Labeling Software},
  year         = {2020--2025},
  howpublished = {\url{https://github.com/HumanSignal/label-studio}},
  note         = {Open source software available from \url{https://github.com/HumanSignal/label-studio}. Accessed: 2026-05-04}
}

@article{liu2022fineaction,
  title={Fineaction: A fine-grained video dataset for temporal action localization},
  author={Liu, Yi and Wang, Limin and Wang, Yali and Ma, Xiao and Qiao, Yu},
  journal={IEEE transactions on image processing},
  volume={31},
  pages={6937--6950},
  year={2022},
  publisher={IEEE}
}

@misc{openai2026gpt54systemcard,
  author       = {{OpenAI}},
  title        = {{GPT-5.4 Thinking System Card}},
  year         = {2026},
  howpublished = {\url{https://openai.com/index/gpt-5-4-thinking-system-card/}},
  note         = {Accessed: 2026-05-04}
}

@misc{google2026gemini31pro,
  author       = {{Google DeepMind}},
  title        = {{Gemini 3.1 Pro}: Model Card},
  year         = {2026},
  howpublished = {\url{https://deepmind.google/models/model-cards/gemini-3-1-pro/}},
  note         = {Accessed: 2026-05-04}
}

@article{hurst2024gpt,
  title={Gpt-4o system card},
  author={Hurst, Aaron and Lerer, Adam and Goucher, Adam P and Perelman, Adam and Ramesh, Aditya and Clark, Aidan and Ostrow, AJ and Welihinda, Akila and Hayes, Alan and Radford, Alec and others},
  journal={arXiv preprint arXiv:2410.21276},
  year={2024}
}

@article{wang2025internvl3,
  title={Internvl3. 5: Advancing open-source multimodal models in versatility, reasoning, and efficiency},
  author={Wang, Weiyun and Gao, Zhangwei and Gu, Lixin and Pu, Hengjun and Cui, Long and Wei, Xingguang and Liu, Zhaoyang and Jing, Linglin and Ye, Shenglong and Shao, Jie and others},
  journal={arXiv preprint arXiv:2508.18265},
  year={2025}
}

@article{bai2025qwen3,
  title={Qwen3-vl technical report},
  author={Bai, Shuai and Cai, Yuxuan and Chen, Ruizhe and Chen, Keqin and Chen, Xionghui and Cheng, Zesen and Deng, Lianghao and Ding, Wei and Gao, Chang and Ge, Chunjiang and others},
  journal={arXiv preprint arXiv:2511.21631},
  year={2025}
}

@article{zhang2025timelens,
  title={TimeLens: Rethinking Video Temporal Grounding with Multimodal LLMs},
  author={Zhang, Jun and Wang, Teng and Ge, Yuying and Ge, Yixiao and Li, Xinhao and Shan, Ying and Wang, Limin},
  journal={arXiv preprint arXiv:2512.14698},
  year={2025}
}

@inproceedings{liuniversal,
  title={Universal Video Temporal Grounding with Generative Multi-modal Large Language Models},
  author={Li, Zeqian and Di, Shangzhe and Zhai, Zhonghua and Huang, Weilin and Wang, Yanfeng and Xie, Weidi},
  booktitle={The Thirty-ninth Annual Conference on Neural Information Processing Systems},
  year={2025}
}

@article{guo2024trace,
  title={Trace: Temporal grounding video llm via causal event modeling},
  author={Guo, Yongxin and Liu, Jingyu and Li, Mingda and Liu, Qingbin and Chen, Xi and Tang, Xiaoying},
  journal={arXiv preprint arXiv:2410.05643},
  year={2024}
}

@inproceedings{wang-etal-2025-grounded,
    title = "Grounded-{V}ideo{LLM}: Sharpening Fine-grained Temporal Grounding in Video Large Language Models",
    author = "Wang, Haibo  and
      Xu, Zhiyang  and
      Cheng, Yu  and
      Diao, Shizhe  and
      Zhou, Yufan  and
      Cao, Yixin  and
      Wang, Qifan  and
      Ge, Weifeng  and
      Huang, Lifu",
    booktitle = "Findings of the Association for Computational Linguistics: EMNLP 2025",
    year = "2025",
    pages = "959--975",
}

@inproceedings{huang2024vtimellm,
  title={Vtimellm: Empower llm to grasp video moments},
  author={Huang, Bin and Wang, Xin and Chen, Hong and Song, Zihan and Zhu, Wenwu},
  booktitle={Proceedings of the IEEE/CVF Conference on Computer Vision and Pattern Recognition},
  pages={14271--14280},
  year={2024}
}

@inproceedings{tang2025sim,
  title={Sim-DETR: Unlock DETR for Temporal Sentence Grounding},
  author={Tang, Jiajin and Wei, Zhengxuan and Zhu, Yuchen and Shi, Cheng and Li, Guanbin and Lin, Liang and Yang, Sibei},
  booktitle={Proceedings of the IEEE/CVF International Conference on Computer Vision},
  pages={22760--22771},
  year={2025}
}

@inproceedings{xiao2024bridging,
  title={Bridging the gap: A unified video comprehension framework for moment retrieval and highlight detection},
  author={Xiao, Yicheng and Luo, Zhuoyan and Liu, Yong and Ma, Yue and Bian, Hengwei and Ji, Yatai and Yang, Yujiu and Li, Xiu},
  booktitle={Proceedings of the IEEE/CVF conference on computer vision and pattern recognition},
  pages={18709--18719},
  year={2024}
}

@article{moon2023correlation,
  title={Correlation-guided query-dependency calibration for video temporal grounding},
  author={Moon, WonJun and Hyun, Sangeek and Lee, SuBeen and Heo, Jae-Pil},
  journal={arXiv preprint arXiv:2311.08835},
  year={2023}
}

@inproceedings{sun2024tr,
  title={Tr-detr: Task-reciprocal transformer for joint moment retrieval and highlight detection},
  author={Sun, Hao and Zhou, Mingyao and Chen, Wenjing and Xie, Wei},
  booktitle={Proceedings of the AAAI Conference on Artificial Intelligence},
  volume={38},
  number={5},
  pages={4998--5007},
  year={2024}
}

@inproceedings{huang2024semantic,
  title={Semantic fusion augmentation and semantic boundary detection: A novel approach to multi-target video moment retrieval},
  author={Huang, Cheng and Wu, Yi-Lun and Shuai, Hong-Han and Huang, Ching-Chun},
  booktitle={Proceedings of the IEEE/CVF Winter Conference on Applications of Computer Vision},
  pages={6783--6792},
  year={2024}
}

@inproceedings{jang2023knowing,
  title={Knowing where to focus: Event-aware transformer for video grounding},
  author={Jang, Jinhyun and Park, Jungin and Kim, Jin and Kwon, Hyeongjun and Sohn, Kwanghoon},
  booktitle={Proceedings of the IEEE/CVF International Conference on Computer Vision},
  pages={13846--13856},
  year={2023}
}

@inproceedings{nishimura2024lighthouse,
  title={Lighthouse: A user-friendly library for reproducible video moment retrieval and highlight detection},
  author={Nishimura, Taichi and Nakada, Shota and Munakata, Hokuto and Komatsu, Tatsuya},
  booktitle={Proceedings of the 2024 Conference on Empirical Methods in Natural Language Processing: System Demonstrations},
  pages={53--60},
  year={2024}
}

@inproceedings{moon2023query,
  title={Query-dependent video representation for moment retrieval and highlight detection},
  author={Moon, WonJun and Hyun, Sangeek and Park, SangUk and Park, Dongchan and Heo, Jae-Pil},
  booktitle={Proceedings of the IEEE/CVF conference on computer vision and pattern recognition},
  pages={23023--23033},
  year={2023}
}

@techreport{nvidia2025cosmosreason2,
    author      = {{NVIDIA}},
    title       = {{Cosmos-Reason2}: A Multimodal Model for Physical and Commonsense Reasoning},
    institution = {NVIDIA},
    year        = {2025},
    type        = {Technical Report},
    url         = {https://huggingface.co/nvidia/Cosmos-Reason2-8B},
    note        = {Model card. Accessed: 2026-05-04}
}

@inproceedings{pavlakos2019expressive,
  title={Expressive body capture: 3d hands, face, and body from a single image},
  author={Pavlakos, Georgios and Choutas, Vasileios and Ghorbani, Nima and Bolkart, Timo and Osman, Ahmed AA and Tzionas, Dimitrios and Black, Michael J},
  booktitle={Proceedings of the IEEE/CVF conference on computer vision and pattern recognition},
  pages={10975--10985},
  year={2019}
}

@article{lin2026graphicontact,
  title={GraphiContact: Pose-aware Human-Scene Robust Contact Perception for Interactive Systems},
  author={Lin, Xiaojian and Shen, Yaomin and Ma, Junyuan and Sun, Yujie and Bu, Chengqing and Zhang, Wenxin and Zhang, Zongzheng and Fei, Hao and Jin, Lei and Zhao, Hao},
  journal={arXiv preprint arXiv:2603.20310},
  year={2026}
}

@inproceedings{caoone,
  title={When One Moment Isn't Enough: Multi-Moment Retrieval with Cross-Moment Interactions},
  author={Cao, Zhuo and Du, Heming and Zhang, Bingqing and Yu, Xin and Li, Xue and Wang, Sen},
  booktitle={The Thirty-ninth Annual Conference on Neural Information Processing Systems},
  year={2025}
}

@inproceedings{grauman2022ego4d,
  title={Ego4d: Around the world in 3,000 hours of egocentric video},
  author={Grauman, Kristen and Westbury, Andrew and Byrne, Eugene and Chavis, Zachary and Furnari, Antonino and Girdhar, Rohit and Hamburger, Jackson and Jiang, Hao and Liu, Miao and Liu, Xingyu and others},
  booktitle={Proceedings of the IEEE/CVF conference on computer vision and pattern recognition},
  pages={18995--19012},
  year={2022}
}

@inproceedings{soldan2022mad,
  title={Mad: A scalable dataset for language grounding in videos from movie audio descriptions},
  author={Soldan, Mattia and Pardo, Alejandro and Alc{\'a}zar, Juan Le{\'o}n and Caba, Fabian and Zhao, Chen and Giancola, Silvio and Ghanem, Bernard},
  booktitle={Proceedings of the IEEE/CVF Conference on Computer Vision and Pattern Recognition},
  pages={5026--5035},
  year={2022}
}

@article{lei2021detecting,
  title={Detecting moments and highlights in videos via natural language queries},
  author={Lei, Jie and Berg, Tamara L and Bansal, Mohit},
  journal={Advances in Neural Information Processing Systems},
  volume={34},
  pages={11846--11858},
  year={2021}
}

@inproceedings{lei2020tvr,
  title={Tvr: A large-scale dataset for video-subtitle moment retrieval},
  author={Lei, Jie and Yu, Licheng and Berg, Tamara L and Bansal, Mohit},
  booktitle={European Conference on Computer Vision},
  pages={447--463},
  year={2020},
  organization={Springer}
}

@inproceedings{krishna2017dense,
  title={Dense-captioning events in videos},
  author={Krishna, Ranjay and Hata, Kenji and Ren, Frederic and Fei-Fei, Li and Carlos Niebles, Juan},
  booktitle={Proceedings of the IEEE international conference on computer vision},
  pages={706--715},
  year={2017}
}

@inproceedings{anne2017localizing,
  title={Localizing moments in video with natural language},
  author={Anne Hendricks, Lisa and Wang, Oliver and Shechtman, Eli and Sivic, Josef and Darrell, Trevor and Russell, Bryan},
  booktitle={Proceedings of the IEEE international conference on computer vision},
  pages={5803--5812},
  year={2017}
}

@inproceedings{gao2017tall,
  title={Tall: Temporal activity localization via language query},
  author={Gao, Jiyang and Sun, Chen and Yang, Zhenheng and Nevatia, Ram},
  booktitle={Proceedings of the IEEE international conference on computer vision},
  pages={5267--5275},
  year={2017}
}

@inproceedings{shinbodycontact4d,
  title={BodyContact4D: A Multi-view Video Dataset for Understanding Human and Environment Interactions},
  author={Shin, Soyong and Lee, Chaeeun and Chen, Holly and Song, Jyun-Ting and Halilaj, Eni and Kitani, Kris},
  booktitle={Thirteenth International Conference on 3D Vision},
  year={2026}
}

@inproceedings{bhatnagar2022behave,
  title={Behave: Dataset and method for tracking human object interactions},
  author={Bhatnagar, Bharat Lal and Xie, Xianghui and Petrov, Ilya A and Sminchisescu, Cristian and Theobalt, Christian and Pons-Moll, Gerard},
  booktitle={Proceedings of the IEEE/CVF conference on computer vision and pattern recognition},
  pages={15935--15946},
  year={2022}
}

@article{gupta2015visual,
  title={Visual semantic role labeling},
  author={Gupta, Saurabh and Malik, Jitendra},
  journal={arXiv preprint arXiv:1505.04474},
  year={2015}
}

@article{comanici2025gemini,
  title={Gemini 2.5: Pushing the frontier with advanced reasoning, multimodality, long context, and next generation agentic capabilities},
  author={Comanici, Gheorghe and Bieber, Eric and Schaekermann, Mike and Pasupat, Ice and Sachdeva, Noveen and Dhillon, Inderjit and Blistein, Marcel and Ram, Ori and Zhang, Dan and Rosen, Evan and others},
  journal={arXiv preprint arXiv:2507.06261},
  year={2025}
}

@inproceedings{song2024moviechat,
  title={Moviechat: From dense token to sparse memory for long video understanding},
  author={Song, Enxin and Chai, Wenhao and Wang, Guanhong and Zhang, Yucheng and Zhou, Haoyang and Wu, Feiyang and Chi, Haozhe and Guo, Xun and Ye, Tian and Zhang, Yanting and others},
  booktitle={Proceedings of the IEEE/CVF Conference on Computer Vision and Pattern Recognition},
  pages={18221--18232},
  year={2024}
}

@inproceedings{gu2018ava,
  title={Ava: A video dataset of spatio-temporally localized atomic visual actions},
  author={Gu, Chunhui and Sun, Chen and Ross, David A and Vondrick, Carl and Pantofaru, Caroline and Li, Yeqing and Vijayanarasimhan, Sudheendra and Toderici, George and Ricco, Susanna and Sukthankar, Rahul and others},
  booktitle={Proceedings of the IEEE conference on computer vision and pattern recognition},
  pages={6047--6056},
  year={2018}
}

@inproceedings{bain2020condensed,
  title={Condensed movies: Story based retrieval with contextual embeddings},
  author={Bain, Max and Nagrani, Arsha and Brown, Andrew and Zisserman, Andrew},
  booktitle={Proceedings of the Asian Conference on Computer Vision},
  year={2020}
}

@article{gilardi2002literature,
  title={Literature survey of contact dynamics modelling},
  author={Gilardi, Gianni and Sharf, Inna},
  journal={Mechanism and machine theory},
  volume={37},
  number={10},
  pages={1213--1239},
  year={2002},
  publisher={Elsevier}
}

@inproceedings{lu2025humoto,
  title={HUMOTO: A 4d dataset of mocap human object interactions},
  author={Lu, Jiaxin and Huang, Chun-Hao Paul and Bhattacharya, Uttaran and Huang, Qixing and Zhou, Yi},
  booktitle={Proceedings of the IEEE/CVF International Conference on Computer Vision},
  pages={10886--10897},
  year={2025}
}

@article{khirodkar2024harmony4d,
  title={Harmony4d: A video dataset for in-the-wild close human interactions},
  author={Khirodkar, Rawal and Song, Jyun-Ting and Cao, Jinkun and Luo, Zhengyi and Kitani, Kris},
  journal={Advances in Neural Information Processing Systems},
  volume={37},
  pages={107270--107285},
  year={2024}
}

@inproceedings{xu2024inter,
  title={Inter-x: Towards versatile human-human interaction analysis},
  author={Xu, Liang and Lv, Xintao and Yan, Yichao and Jin, Xin and Wu, Shuwen and Xu, Congsheng and Liu, Yifan and Zhou, Yizhou and Rao, Fengyun and Sheng, Xingdong and others},
  booktitle={Proceedings of the IEEE/CVF conference on computer vision and pattern recognition},
  pages={22260--22271},
  year={2024}
}

@inproceedings{yin2023hi4d,
  title={Hi4d: 4d instance segmentation of close human interaction},
  author={Yin, Yifei and Guo, Chen and Kaufmann, Manuel and Zarate, Juan Jose and Song, Jie and Hilliges, Otmar},
  booktitle={Proceedings of the IEEE/CVF Conference on Computer Vision and Pattern Recognition},
  pages={17016--17027},
  year={2023}
}

@inproceedings{cseke2025pico,
  title={PICO: Reconstructing 3D people in contact with objects},
  author={Cseke, Alp{\'a}r and Tripathi, Shashank and Dwivedi, Sai Kumar and Lakshmipathy, Arjun S and Chatterjee, Agniv and Black, Michael J and Tzionas, Dimitrios},
  booktitle={Proceedings of the Computer Vision and Pattern Recognition Conference},
  pages={1783--1794},
  year={2025}
}

@inproceedings{jiang2023full,
  title={Full-body articulated human-object interaction},
  author={Jiang, Nan and Liu, Tengyu and Cao, Zhexuan and Cui, Jieming and Zhang, Zhiyuan and Chen, Yixin and Wang, He and Zhu, Yixin and Huang, Siyuan},
  booktitle={Proceedings of the IEEE/CVF International Conference on Computer Vision},
  pages={9365--9376},
  year={2023}
}

@inproceedings{xie2022chore,
  title={Chore: Contact, human and object reconstruction from a single rgb image},
  author={Xie, Xianghui and Bhatnagar, Bharat Lal and Pons-Moll, Gerard},
  booktitle={European Conference on Computer Vision},
  pages={125--145},
  year={2022},
  organization={Springer}
}

@article{yang2024egochoir,
  title={Egochoir: Capturing 3d human-object interaction regions from egocentric views},
  author={Yang, Yuhang and Zhai, Wei and Wang, Chengfeng and Yu, Chengjun and Cao, Yang and Zha, Zheng-Jun},
  journal={Advances in Neural Information Processing Systems},
  volume={37},
  pages={54529--54557},
  year={2024}
}

@article{regneri2013grounding,
  title={Grounding action descriptions in videos},
  author={Regneri, Michaela and Rohrbach, Marcus and Wetzel, Dominikus and Thater, Stefan and Schiele, Bernt and Pinkal, Manfred},
  journal={Transactions of the Association for Computational Linguistics},
  volume={1},
  pages={25--36},
  year={2013},
  publisher={MIT Press One Rogers Street, Cambridge, MA 02142-1209, USA journals-info~…}
}

@inproceedings{taheri2020grab,
  title={GRAB: A dataset of whole-body human grasping of objects},
  author={Taheri, Omid and Ghorbani, Nima and Black, Michael J and Tzionas, Dimitrios},
  booktitle={European conference on computer vision},
  pages={581--600},
  year={2020},
  organization={Springer}
}

@inproceedings{hassan2019resolving,
  title={Resolving 3D human pose ambiguities with 3D scene constraints},
  author={Hassan, Mohamed and Choutas, Vasileios and Tzionas, Dimitrios and Black, Michael J},
  booktitle={Proceedings of the IEEE/CVF international conference on computer vision},
  pages={2282--2292},
  year={2019}
}

@inproceedings{huang2022capturing,
  title={Capturing and inferring dense full-body human-scene contact},
  author={Huang, Chun-Hao P and Yi, Hongwei and H{\"o}schle, Markus and Safroshkin, Matvey and Alexiadis, Tsvetelina and Polikovsky, Senya and Scharstein, Daniel and Black, Michael J},
  booktitle={Proceedings of the IEEE/CVF Conference on Computer Vision and Pattern Recognition},
  pages={13274--13285},
  year={2022}
}

@inproceedings{tripathi2023deco,
  title={Deco: Dense estimation of 3d human-scene contact in the wild},
  author={Tripathi, Shashank and Chatterjee, Agniv and Passy, Jean-Claude and Yi, Hongwei and Tzionas, Dimitrios and Black, Michael J},
  booktitle={Proceedings of the IEEE/CVF International Conference on Computer Vision},
  pages={8001--8013},
  year={2023}
}

@inproceedings{chen2023detecting,
  title={Detecting human-object contact in images},
  author={Chen, Yixin and Dwivedi, Sai Kumar and Black, Michael J and Tzionas, Dimitrios},
  booktitle={Proceedings of the IEEE/CVF Conference on Computer Vision and Pattern Recognition},
  pages={17100--17110},
  year={2023}
}
\clearpage

%%%%%%%%%%%%%%%%%%%%%%%%%%%%%%%%%%%%%%%%%%%%%%%%%%%%%%%%%%%%

\newpage
\appendix

\section{Appendix}

\subsection{Broader Impacts}
\label{sec:broader_impacts}

HAPACT broadens video-based physical understanding by treating impact as a human-centric physical cue. While many interaction datasets focus on intentional human actions, impact events often involve external forces, object collisions, and environmental contact. This perspective can support research on how dynamic events affect the human body in visual media, with possible applications in physical event analysis, body-aware video understanding, haptic generation for audiovisual content, and mixed reality experiences. The annotation pipeline may also be adapted to other video domains that require human-centered physical event annotations.

The potential negative impact is limited because HAPACT is built from staged movie clips and does not redistribute the underlying copyrighted footage. However, the dataset contains scenes with violence, falls, collisions, and explosions. Models trained or evaluated on this benchmark should therefore be used with care when transferred beyond the domain, where predictions may be misinterpreted without appropriate context.

\subsection{Video Availability}
\label{sec:video_availability}

HAPACT is constructed from videos collected from the \textit{Movieclips} YouTube channel, following prior movie-based datasets that use the same source~\citep{bain2020condensed, sadhu2021visual}. This source offers relatively stable availability compared to YouTube videos uploaded by individual users. For example, \citet{bain2020condensed} reported that only $0.3\%$ of videos were removed from the \textit{Movieclips} channel between January 1 and September 1, 2020. To support reproducibility, we provide source video identifiers, trimmed timestamps, IMDb identifiers for metadata, annotations, and evaluation scripts, while the underlying movie footage remains owned by the original rights holders.

\subsection{Dataset Construction Details}
\label{sec:construction_details}

\subsubsection{Video Collection and Preprocessing}

We constructed our initial video pool from the \textit{Movieclips} YouTube channel, using videos available as of August $12$, $2025$. To remove promotional outros while avoiding unnecessary trimming, we applied the fixed-duration trimming rule of \citet{bain2020condensed} only when an outro was detected. Specifically, we segmented each video into shots using PySceneDetect~\citep{pyscenedetect} and compared the terminal shot with reference outro clips based on visual similarity. If an outro match was found, we removed the final $10$ seconds for videos uploaded before May $2017$ and the final $30$ seconds for videos uploaded from May $2017$ onward; otherwise, the video was left untrimmed.

\subsubsection{Genre-Based Scoring Prior from 4DX Films}
\label{sec:genre-based scoring}

To construct a video pool rich in physical impact events, we selected candidate movies whose genres were similar to those of films previously screened in 4DX. We compiled a list of 4DX-screened films as of August $7$, $2025$, and computed how frequently each genre and genre pair appeared among them based on IMDb genre metadata. For each candidate movie \(m\), let \(G_m\) denote its set of IMDb genres assigned to \(m\). The single-genre score is defined as
\[
S_{\mathrm{single}}(m)=\frac{1}{|G_m|}\sum_{g\in G_m}p(g),
\]
where \(p(g)\) is the empirical probability of genre \(g\) among the 4DX-screened films. For movies with at least two valid genres, we define the genre-pair score as
\[
S_{\mathrm{pair}}(m)=\frac{1}{|\mathcal{P}_m|}\sum_{(g_i,g_j)\in\mathcal{P}_m}p(g_i,g_j),
\]
where \(\mathcal{P}_m\) denotes the set of genre pairs in \(G_m\), and \(p(g_i,g_j)\) is the empirical probability of the genre pair among the 4DX-screened films. The final genre-based score is defined as
\[
S(m)=S_{\mathrm{single}}(m)+S_{\mathrm{pair}}(m).
\]
Movies with no valid genres were assigned a score of zero, and \(S_{\mathrm{pair}}(m)\) was set to zero when fewer than two valid genres were available. Candidate movies satisfying \(S(m)\geq0.1\) were retained.

\subsubsection{Query-Clip Candidate Generation}

After filtering the candidate movie pool, we generated candidate query-clip pairs for physical impact events with Gemini 2.5 Pro~\citep{comanici2025gemini}. Given each video, the model was prompted to identify second-level temporal segments in which a person experiences a physical impact, and to return each segment with a short text description. The prompt was as follows:

\begin{table*}[h]
\centering
\setlength{\tabcolsep}{8pt}
\renewcommand{\arraystretch}{1.15}
\begin{tabularx}{\linewidth}{>{\raggedright\arraybackslash\setlength{\parindent}{0pt}}X}
\hline
\rowcolor{black!18}
Prompt for Query-Clip Candidate Generation \\
\hline
\addlinespace[0.5em]
Identify segments where a person experiences a physical impact. Consider only sudden and salient physical forces affecting the subject. Exclude light contact, minor bumps, distant events, off-screen events, or events that are not physically relevant to the affected subject. \\
\addlinespace[0.5em]
Answer in the format ``[MM:SS - MM:SS] Description of the impact''. If there are multiple events, return multiple lines. If none, answer ``None''. \\
\addlinespace[0.3em]
\hline
\end{tabularx}
\label{tab:query_clip_prompt}
\end{table*}

The returned descriptions formed the initial impact queries. Since the model outputs second-level timestamps in the \texttt{MM:SS} format, each segment was expanded with a $2.5$-second temporal buffer on both sides. If the buffer exceeded the clip boundaries, the interval was clipped to remain within the original video and aligned to a $0.5$-second step. The resulting intervals were extracted and re-encoded at $24$~FPS to form the query-clip candidate pairs.

\subsubsection{Human-Targeted Impact Filtering}

To remove impact descriptions that do not involve humans, we applied two complementary filters. First, GPT-4o~\citep{hurst2024gpt} was prompted to determine whether each impact physically involves a human, returning a true-or-false label. The prompt counted a description as human-involved when a human was the recipient or direct cause of the impact, or was physically affected by the resulting force:

\begin{table*}[h]
\centering
\setlength{\tabcolsep}{8pt}
\renewcommand{\arraystretch}{1.15}
\begin{tabularx}{\linewidth}{>{\raggedright\arraybackslash\setlength{\parindent}{0pt}}X}
\hline
\rowcolor{black!18}
Prompt for Human-Targeted Impact Filtering \\
\hline
\addlinespace[0.5em]
Determine whether the described impact physically involves a human. Count as True if a human is the recipient of the impact, causes the impact through direct physical action, or is directly affected by the resulting physical force. Do not count cases where a human only observes, hears, or is merely near the impact without being physically affected. \\
\addlinespace[0.5em]
Answer with only ``True'' or ``False''. \\
\addlinespace[0.3em]
\hline
\end{tabularx}
\label{tab:human_filter_prompt}
\end{table*}

Second, we applied a Stanford Parser-based filter~\citep{chen2014fast} to identify descriptions whose affected argument denotes a person. Following the human-subject filtering of \citet{gupta2015visual}, we parsed each description and checked whether the argument associated with the impact predicate matched a list of person-denoting nouns and pronouns. We retained the union of query-clip pairs accepted by the two filters.

\subsubsection{Recall Evaluation of the Candidate Pipeline}
\label{sec:recall_evaluation}

To assess the recall of the candidate generation and filtering steps, we conducted an additional evaluation with authors blinded to the candidate outputs. Pairs of authors annotated all human-targeted impacts across the \textit{Movieclips} videos used in HAPACT construction at one-second resolution and cross-checked their annotations. We then compared these annotations with the candidate results retained after human-targeted impact filtering. The resulting mean recall was 88.99\% (95\% confidence interval: 87.98--90.00\%), indicating that the pipeline captures most human-targeted impacts present in the source videos.

\subsubsection{Manual Revision}

Following human-targeted filtering, author pairs reviewed all query-clip candidates and excluded cases outside the task scope:

\begin{itemize}
    \item \textbf{Impact event affecting a single target person.} As an exception, two-person impacts were retained only when the event involved direct human-human impact.
    \item \textbf{Impact event visible and reliably inferable.} Ambiguous, extremely off-screen, severely blurred, or occluded impacts were excluded.
    \item \textbf{Physically plausible impact event.} Non-physical or supernatural impacts were discarded.
\end{itemize}

A candidate was accepted only when both authors judged it to satisfy these criteria. When a valid candidate instance could be obtained by editing the query, one author proposed the revision and the other reviewed it for approval. If both authors proposed different revisions for the same instance, the final query was selected by consensus.

\subsection{Dataset Annotation Details}
\label{sec:annotation_details}

\subsubsection{Annotator Selection and Distribution}
\label{sec:annotator_selection}

Annotators were selected separately for each annotation task after a training phase. For each task, applicants annotated a screening set and their results were evaluated against author-annotated pseudo ground truth. Among $52$ applicants, $28$ annotators were selected. The selected annotators included $14$ female annotators, with $7$ graduate students and $21$ undergraduate students. Annotators were compensated at a rate of $\$0.12$ per video-query pair for temporal annotation and $\$1.50$ per video-query pair for spatial annotation.

The annotation protocol was approved by the KAIST Institutional Review Board (IRB No. KAISTIRB-2025-235). Annotators were informed before participation that the task involved viewing movie clips with physical impact events, including violence, falls, collisions, and explosions, which could cause discomfort or fatigue.

\subsubsection{Annotation Instructions}
\label{sec:annotation_instructions}

Annotators were provided with task-specific instruction materials before each annotation stage. They were trained on a definition of physical impact based on~\citet{gilardi2002literature}. Following this definition, annotators identified each occurrence of the queried impact, marking its temporal extent and the body region affected in each impact frame. The core instructions from the materials are reproduced below.

\begin{table*}[!h]
\centering
\setlength{\tabcolsep}{8pt}
\renewcommand{\arraystretch}{1.15}
\begin{tabularx}{\linewidth}{>{\raggedright\arraybackslash\setlength{\parindent}{0pt}}X}
\hline
\rowcolor{black!18}
Instructions for Temporal Impact Annotation \\
\hline
\addlinespace[0.5em]
You are given a movie clip and an impact description. \newline

Your task is to mark the start and end frames of the physical impact described in the video. Review the impact description and identify when the impact is applied to the person. Mark only the impact timing itself, excluding the approach before the impact and the aftermath after the impact. If the described impact occurs multiple times, mark each occurrence separately. \newline

If the queried impact is not visible or cannot be reliably inferred, leave the comment and skip. \\
\addlinespace[0.3em]
\hline
\end{tabularx}
\label{tab:temporal_annotation_instruction}
\end{table*}

\begin{table*}[!h]
\centering
\setlength{\tabcolsep}{8pt}
\renewcommand{\arraystretch}{1.15}
\begin{tabularx}{\linewidth}{>{\raggedright\arraybackslash\setlength{\parindent}{0pt}}X}
\hline
\rowcolor{black!18}
Instructions for Spatial On-body Impact Annotation \\
\hline
\addlinespace[0.5em]
You are given a movie clip, an impact description, and the temporal interval(s) for the impact. \newline

Carefully review the clip, the impact description, and the provided interval(s). For every frame within the interval(s), identify which part of the human body is affected by the impact and mark the affected region using the web-based annotation tool. All frames inside the provided interval(s) must be annotated. For human-human impacts, use the grammatical subject of the impact description as the target person, and annotate only the body region of that subject. \newline

If the impact is not perceived on the subject in a frame, or if the affected body region cannot be inferred, briefly explain the reason in the comment and remove that frame from the interval. \\
\addlinespace[0.3em]
\hline
\end{tabularx}
\label{tab:spatial_annotation_instruction}
\end{table*}

\subsubsection{Temporal Annotation Aggregation}
\label{sec:temporal_annotation_aggregation}

For temporal annotation, each video-query pair was annotated by three annotators. To aggregate them, each annotated interval was first converted into a frame-level binary label. The final temporal intervals were obtained by majority voting, where a frame was retained if at least two of the three annotators labeled it as containing the queried event. Ambiguous cases after aggregation were resolved by author consensus. These included cases with low agreement among annotators, measured by a mean pairwise tIoU of at most $0.5$, and cases where majority voting collapsed the annotated intervals into fewer moments.

\begin{figure*}[t]
    \centering
    \includegraphics[width=\textwidth]{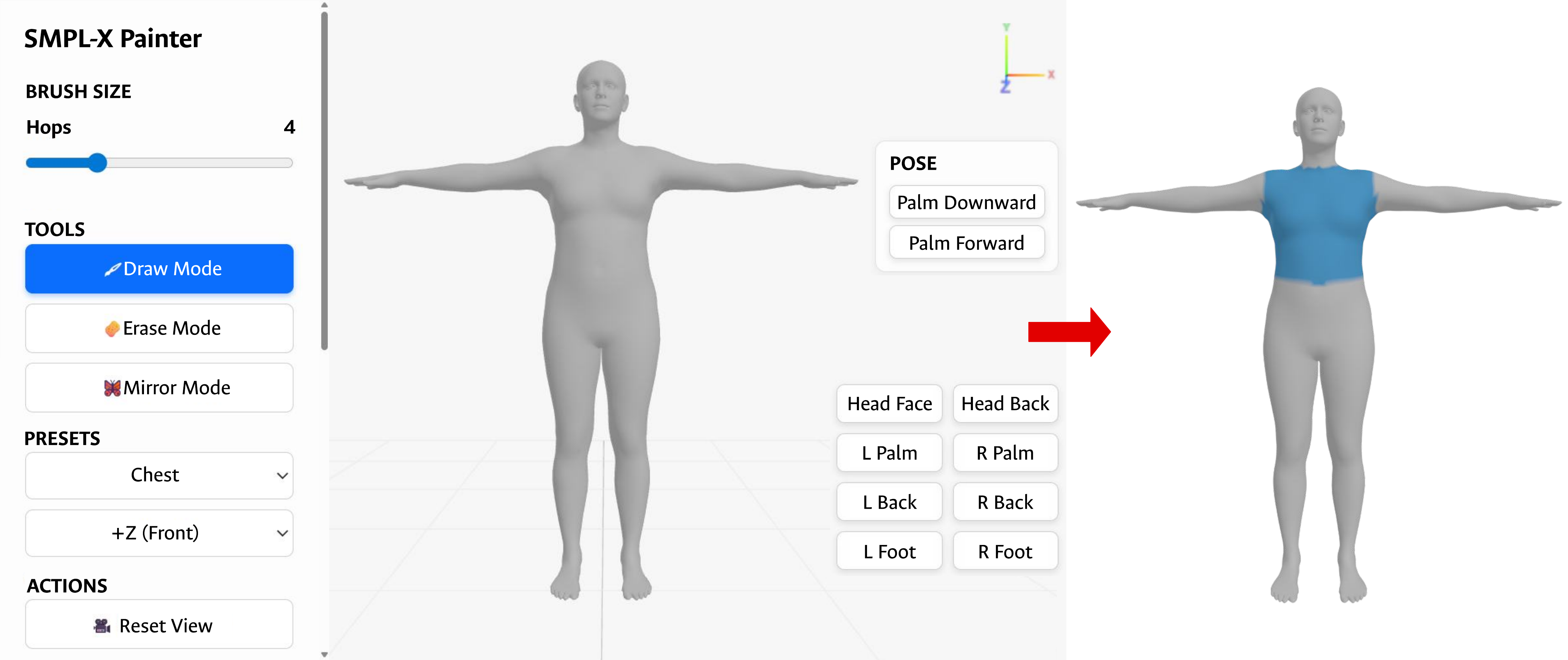}
    \caption{\textbf{Spatial annotation interface.} Annotators mark affected body regions on an SMPL-X mesh using draw and erase modes, with controls for mesh navigation, brush size, and shortcut views.}
    \label{fig:spatial_annotation_interface}
\end{figure*}

\subsubsection{Spatial Annotation Interface}
\label{sec:spatial_annotation_interface}

For spatial annotation, we built a dense on-body 3D annotation interface inspired by prior 3D human contact annotation datasets~\citep{tripathi2023deco,yang2024lemon}. The interface centers on an SMPL-X mesh that can be translated, rotated, and zoomed. The left panel provides controls for switching between draw and erase modes, as well as a slider for adjusting the brush size. In draw mode, annotators mark affected regions by clicking and dragging on the mesh, while in erase mode, they remove marked regions. The lower-right panel provides shortcut buttons for specific views, including the palms, feet, face, and back of the head. To reduce ambiguity caused by differences between the canonical and observed poses, the annotation interface provided multiple canonical T-pose variants. The interface was deployed as a web-based tool to be used alongside the HumanSignal platform. Annotators reviewed the video, query, and temporal intervals for each assigned data instance in HumanSignal, marked the affected body regions for each frame using the interface, and copied the resulting vertex indices back to HumanSignal. The interface design is shown in Fig.~\ref{fig:spatial_annotation_interface}. Impact intervals for which annotators could not infer the affected body region with sufficient confidence were excluded from the final dataset.

\subsubsection{Quality Control and Assessment} 
\label{sec:quality_control_assessment}

To maintain annotation quality, we applied two quality-control procedures across both annotation stages. First, as described in Section~\ref{sec:annotator_selection}, annotators were selected through a pre-screening task in which their annotations were evaluated against author-annotated references. Second, all submitted annotations were reviewed by meta-annotators, and low-quality cases were returned for re-annotation with explanatory feedback.

Furthermore, annotation quality was evaluated using both inter-annotator agreement and reference-based validation, following the protocol of \citet{tripathi2023deco}. For temporal annotation, annotators labeled a shared set of $10$ impact moments from $8$ movies to measure agreement, yielding a Fleiss' $\kappa$ of $0.46$. Fleiss' $\kappa$ ranges from $-1$ to $1$, with $-1$, $0$, and $1$ indicating complete disagreement, chance-level agreement, and complete agreement, respectively. A value of $0.46$ indicates moderate agreement. To further validate boundary accuracy on existing data, a curated set of $100$ physical-interaction instances from FineAction~\citep{liu2022fineaction} was annotated under the same protocol. The resulting mean temporal IoU was $0.70$, indicating reliable temporal boundary annotation.

For spatial annotation, annotators labeled $20$ shared frames from $7$ movies to measure agreement. Fleiss' $\kappa$ was computed over all SMPL-X vertices, yielding $0.53$ at the vertex level. We additionally measured body-part-level agreement by aggregating vertices according to the official SMPL-X body segmentation provided by Meshcapade\footnote{Available at \url{https://github.com/Meshcapade/wiki/tree/main/assets/SMPL_body_segmentation/smplx}}, which yielded a Fleiss' $\kappa$ of $0.61$. These results show stronger consistency at the body-part level, with the vertex-level score comparable to those reported in representative human-annotated physical-interaction datasets~\citep{bansal2026towards, tripathi2023deco}.

To assess whether annotators could consistently produce labels from images with occlusions and diverse body poses, we evaluated their annotations against sensor-derived 3D references on 100 samples. We followed prior work~\citep{tripathi2023deco}, which also uses canonical-mesh-based annotations. The data were curated from PROX~\citep{hassan2019resolving} and RICH~\citep{huang2022capturing}, and annotated following the human-scene contact task protocol to evaluate consistency with external mesh-based reference labels. For these samples, we manually created text queries to constrain which body region should be marked. The resulting mIoU with the reference labels was $0.42$, comparable to the value reported by \citet{tripathi2023deco}.

\begin{figure*}[t]
\centering
\includegraphics[width=0.49\textwidth]{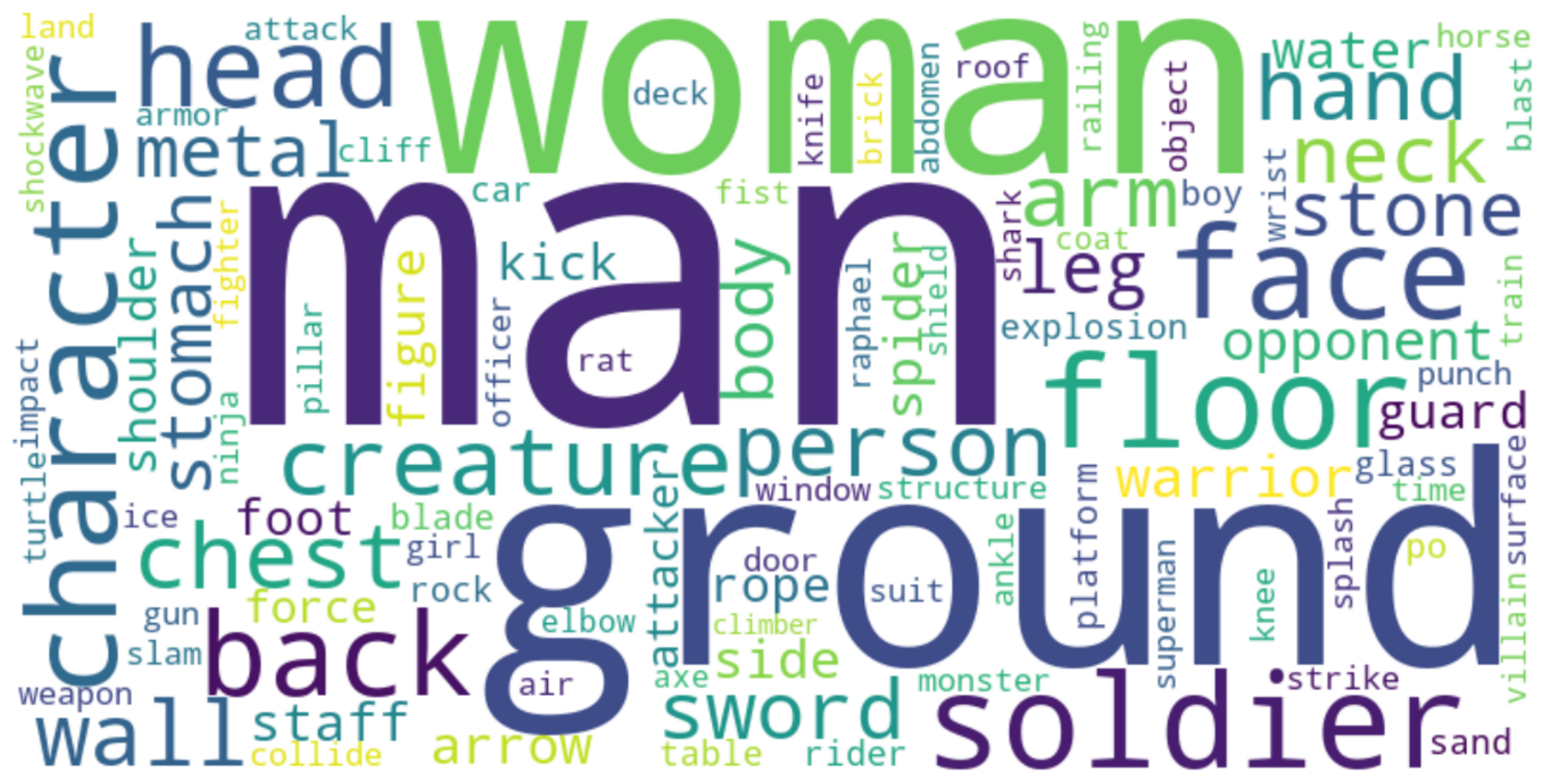}
\includegraphics[width=0.49\textwidth]{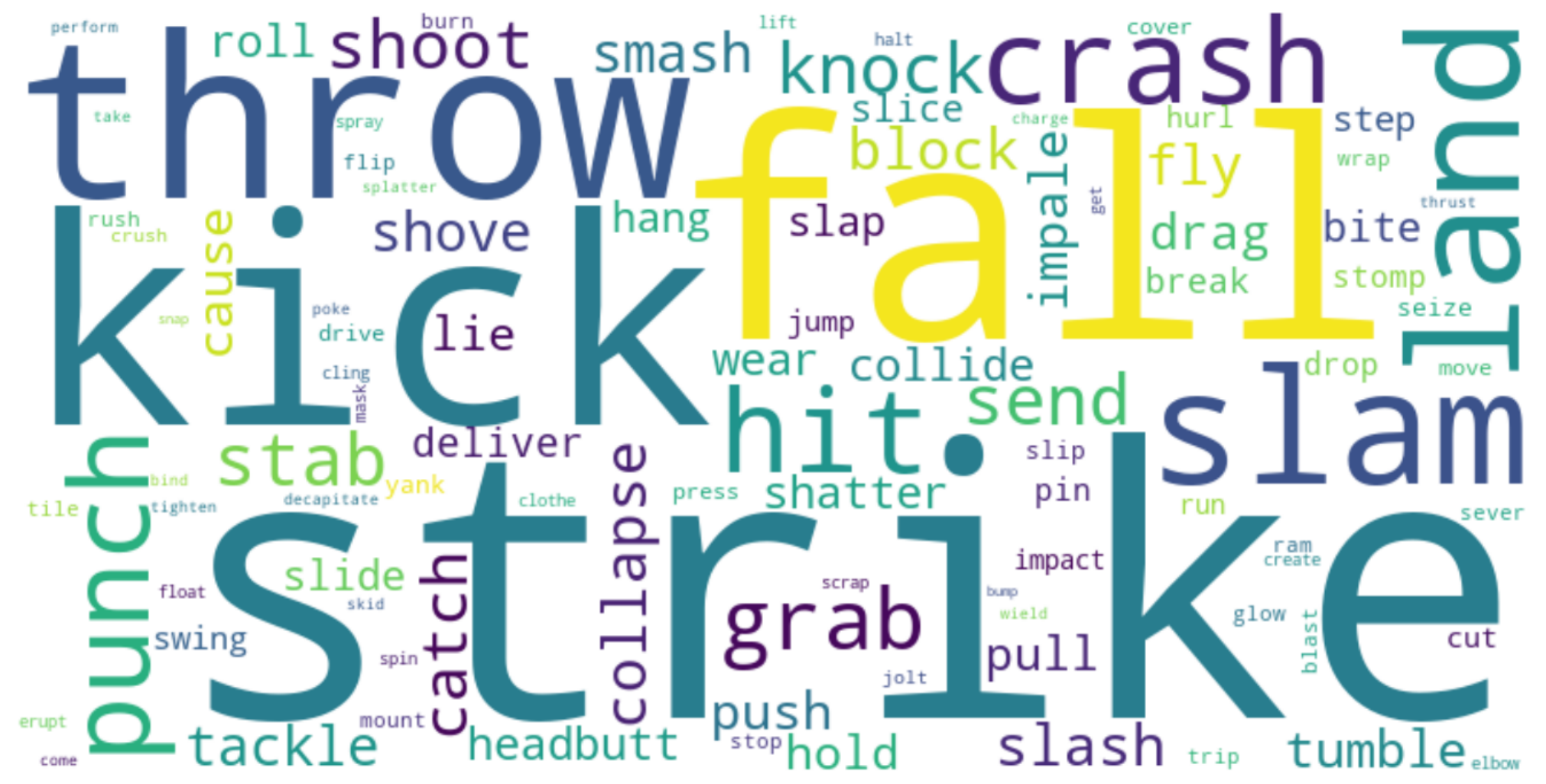}
\caption{Distribution of nouns (left) and verbs (right) in the descriptions.}
\label{fig:voca}
\end{figure*}

\begin{table*}[t]
\centering
\begin{tabular}{lrrrr}
\toprule[1.5pt]
\textbf{} & \textbf{Total} & \textbf{Train} & \textbf{Test} & \textbf{Val} \\
 \midrule
\multicolumn{5}{l}{\color{gray}\textit{Video Clip Statistics}} \\
Number of Movies                      & 212     & 148     & 32     & 32     \\
Number of Clips                       & 2,884   & 2,019   & 433    & 432    \\
Mean Clip Duration             & 6.49s   & 6.49s   & 6.44s  & 6.55s  \\
\midrule
\multicolumn{5}{l}{\color{gray}\textit{Query Statistics}} \\
Total Word Count               & 32,039  & 22,395  & 4,920  & 4,724  \\
Mean Word Count per Query      & 11.1    & 11.1    & 11.4   & 10.9   \\
\midrule
\multicolumn{5}{l}{\color{gray}\textit{Moment Length Statistics}} \\
Number of Moments              & 4,242  & 2,956  & 649   & 637   \\
Mean Frames Count per Moment   & 4.80    & 4.83    & 4.84   & 4.65   \\
\quad Short (1--2 frames)      & 45.33\% & 45.53\% & 44.68\% & 45.05\% \\
\quad Middle (3--6 frames)     & 34.35\% & 34.10\% & 34.51\% & 35.32\% \\
\quad Long (7+ frames)         & 20.32\% & 20.37\% & 20.80\% & 19.62\% \\
\bottomrule[1.5pt]
\end{tabular}
\caption{Statistics on splits of HAPACT.}
\label{tab:appendix_stats}
\end{table*}

\subsection{More Dataset Statistics}
\label{sec:statistics_details}

\subsubsection{Overall Dataset Statistics}
\label{sec:statistics_overall}

Figure~\ref{fig:voca} shows the distribution of frequent nouns and verbs extracted from query descriptions. Frequent nouns such as ``man'' and ``woman'' reflect the human-centric nature of the dataset. Verbs such as ``strike'' and ``kick'' indicate that the dataset primarily deals with direct physical impact. 

\clearpage
\newpage

\begin{table*}[p]
\centering
\begin{tabular}{crrrr}
\toprule[1.5pt]
\textbf{Moments per Query} & \textbf{Total} & \textbf{Train} & \textbf{Test} & \textbf{Val} \\
 \midrule
1  & 1,900 & 1,355 & 272 & 273 \\
2  & 753   & 507   & 119 & 127 \\
3+  & 231   & 157   & 42  & 32  \\
\midrule
Total & 2,884 & 2,019 & 433 & 432 \\
\bottomrule[1.5pt]
\end{tabular}
\caption{Distribution of the number of moments per query for each split.}
\label{tab:moment_distribution}
\end{table*}

\begin{figure*}[p]
\centering
\includegraphics[width=\linewidth]{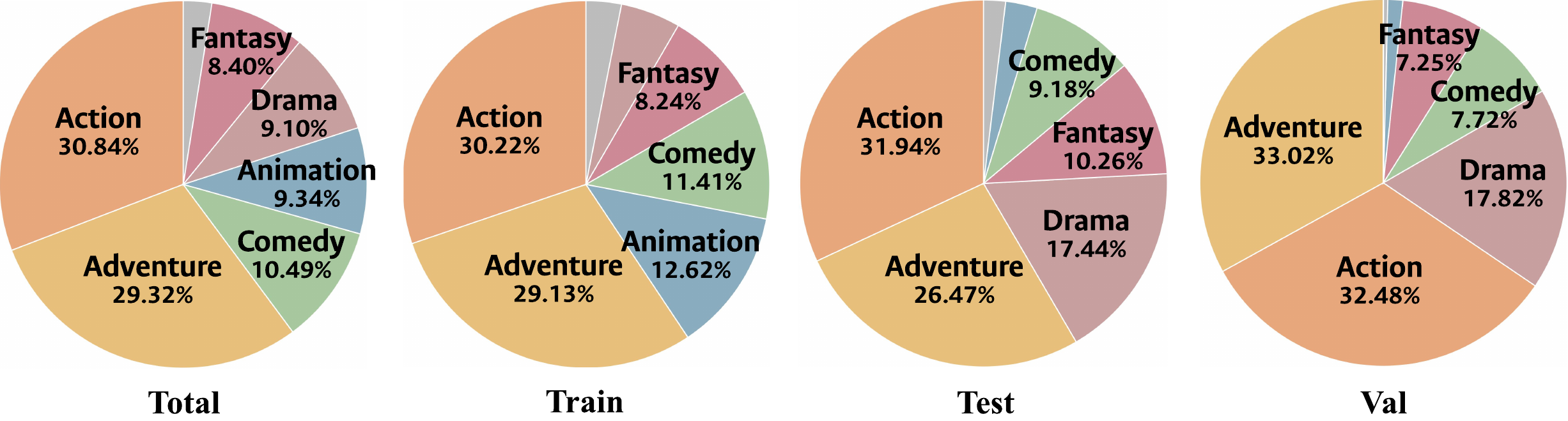}
\caption{\textbf{Distribution of movie genres for each split.} This shows that action and adventure comprise a similar proportion of each split.}
\label{fig:genre}
\end{figure*}

\begin{figure*}[p]
\centering
\includegraphics[width=0.24\textwidth]{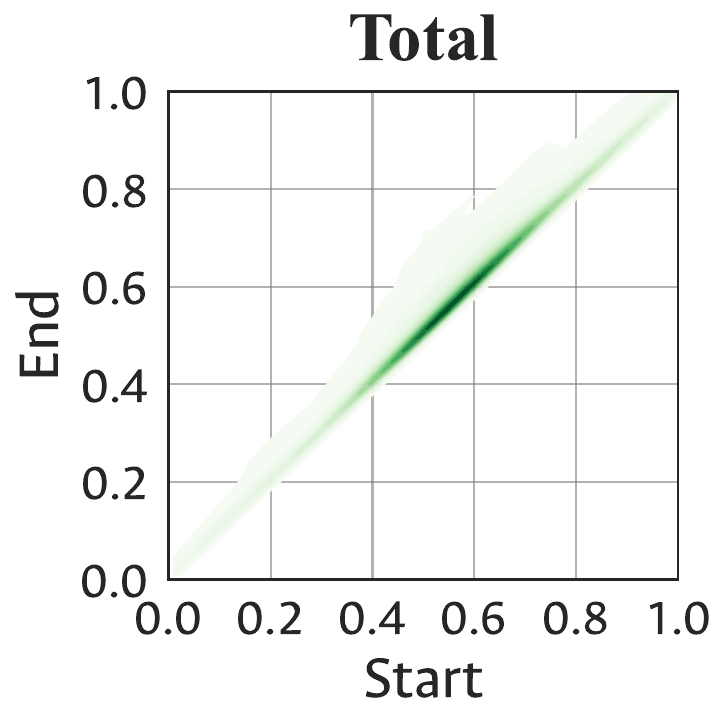}
\includegraphics[width=0.24\textwidth]{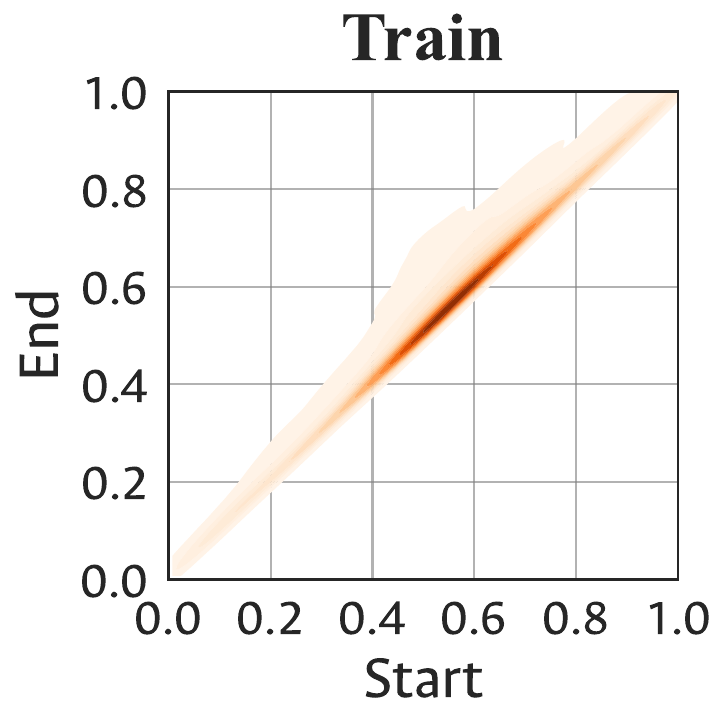}
\includegraphics[width=0.24\textwidth]{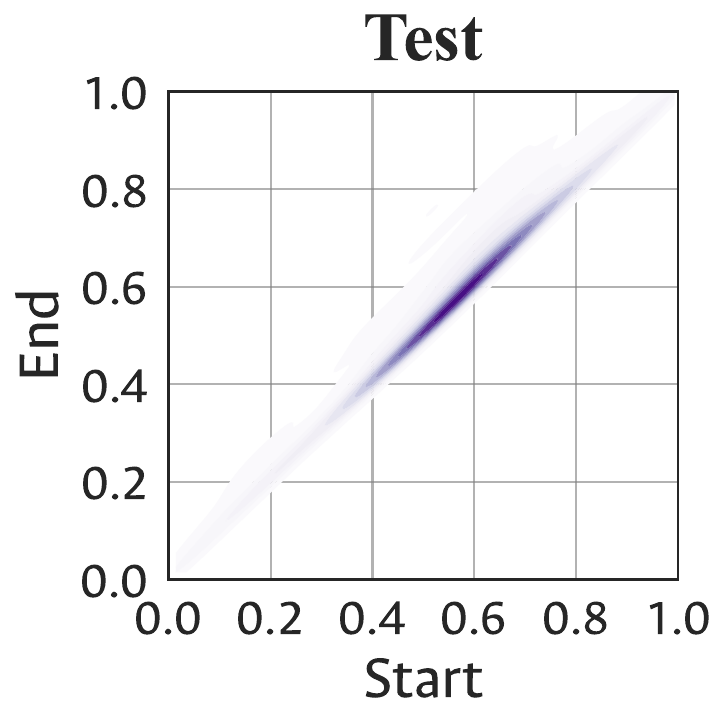}
\includegraphics[width=0.24\textwidth]{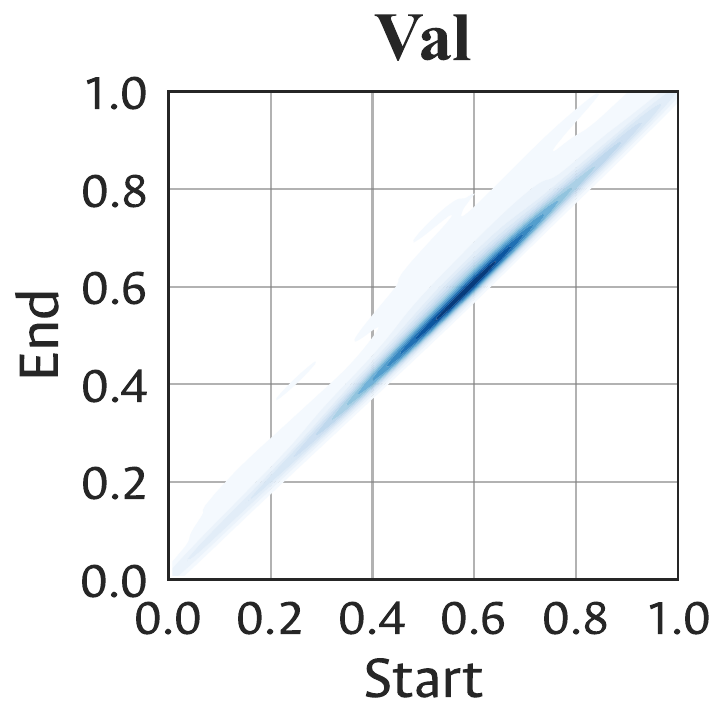}
\caption{\textbf{Distribution of start-end positions for each split.} Start and end positions of impact moments, normalized from 0 to 1 relative to the total video length. The distribution of moment positions is comparable across all splits.}
\label{fig:position}
\end{figure*}

\begin{figure*}[p]
\centering
\includegraphics[width=\linewidth]{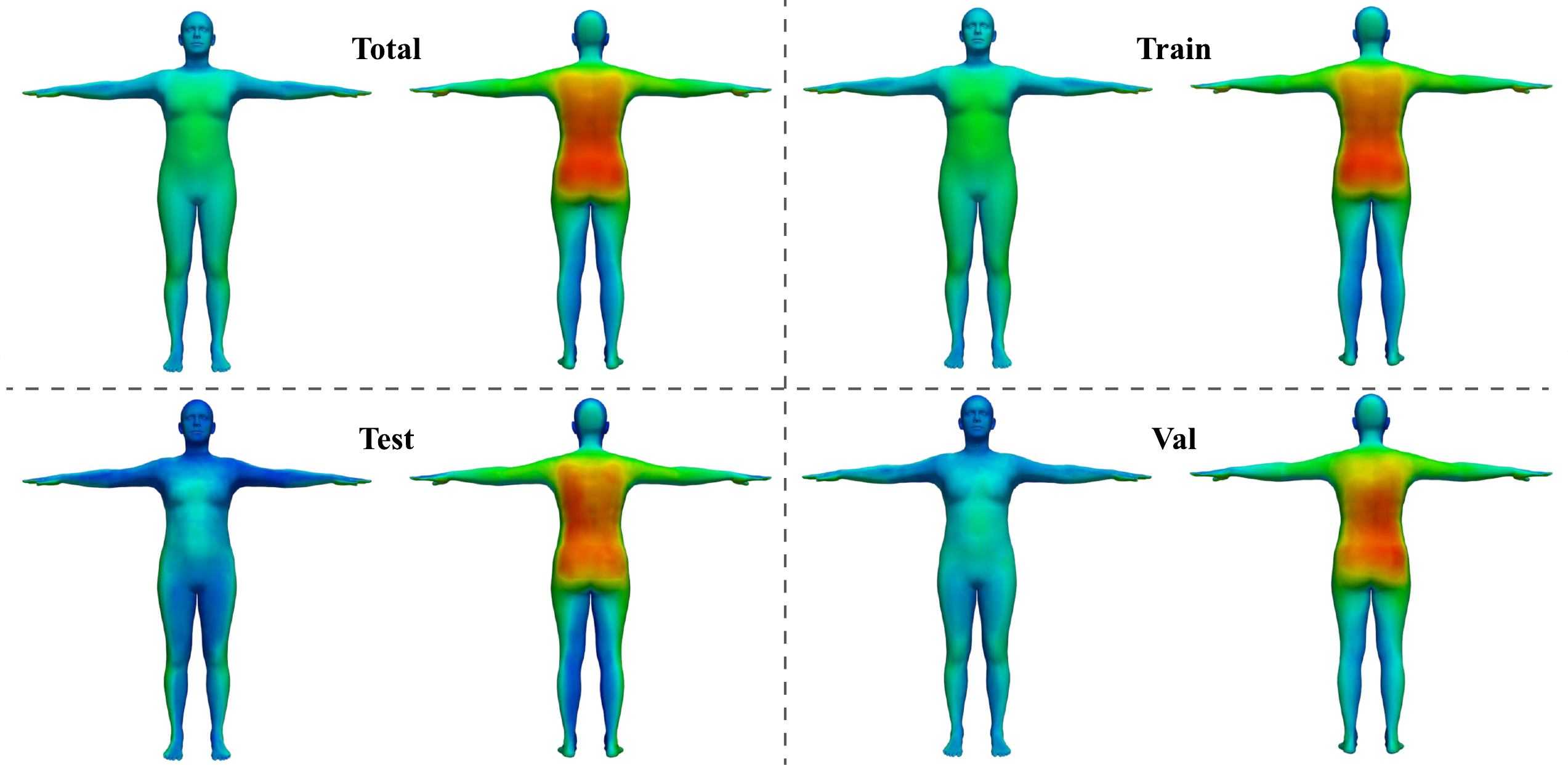}
\caption{\textbf{Distribution of impact occurrences on the body mesh for each split.} 1 indicates a high probability of contact and 0 indicates a low probability. This indicates that the affected body vertices were distributed similarly across the splits.}
\label{fig:heatmap}
\end{figure*}

\clearpage
\newpage

\subsubsection{Split Statistics}
\label{sec:statistics_split}

Table~\ref{tab:appendix_stats} shows the statistics for each split, across the clip, query, and moment distributions. This confirms that the splits are structured to maintain consistent distributions of clip duration, number of moments, and moment length. Table~\ref{tab:moment_distribution} shows the distribution of the number of moments per query. The dataset includes queries with an average of $1.47$ moments, as well as a significant proportion of queries with two or more moments.

\begin{figure*}[t]
\centering
\includegraphics[width=\textwidth]{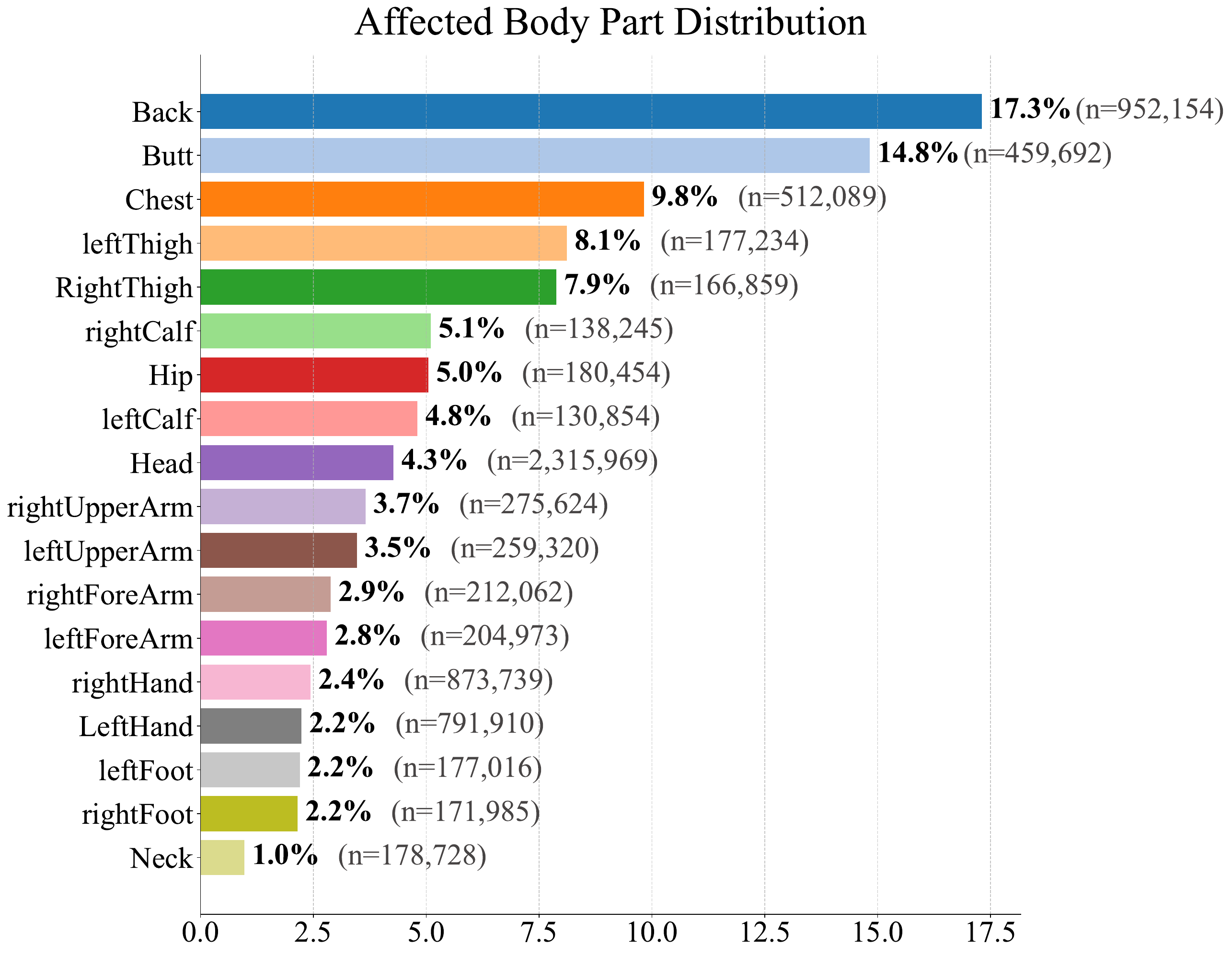}
\caption{\textbf{Distribution of affected body parts weighted by face-level intensity across the SMPL-X mesh.} Each bar represents the face-level intensity score (\%) for each body part.}
\label{fig:body_contact_dist}
\end{figure*}

\subsection{Evaluation Details}
\label{sec:evaluation_details}

\subsubsection{General Evaluation Settings}

Evaluation was conducted on a server equipped with $4$ NVIDIA RTX A6000 GPUs, each with $48$GB of memory. Benchmark models were evaluated on the HAPACT test split, while models requiring training or fine-tuning were trained on the training split and selected using the validation split.

\subsubsection{Temporal Localization Evaluation}
\label{sec:temporal_evaluation_details}

For retrieval-based baselines, CLIP features were extracted at a sampling interval of $0.05$ seconds to provide dense frame coverage for each clip. Both visual and query features were obtained following the CLIP feature extraction pipeline provided by Lighthouse~\citep{nishimura2024lighthouse}. Evaluation was conducted using the same framework and configurations, which provide a reproducible codebase for video moment retrieval. Since Sim-DETR and SFABD are not included in Lighthouse, they were evaluated using their official repositories. Models were initialized from weights pretrained on QVHighlights and fine-tuned for $200$ epochs with a learning rate of $1\times10^{-5}$. For methods using saliency supervision, the saliency loss weight was set to $1.0$.

\begin{table*}[!p]
\centering
\small
\setlength{\tabcolsep}{6pt}
\renewcommand{\arraystretch}{1.15}

\caption{\textbf{Prompts for Generation-based Temporal Localization Baselines.} Each prompt follows the model's native input convention while preserving the same impact-localization constraints.}
\begin{tabularx}{\linewidth}{
>{\raggedright\arraybackslash}p{0.14\linewidth}
>{\raggedright\arraybackslash\setlength{\parindent}{0pt}}X}
\hline
\rowcolor{black!18}
Model & Prompt \\
\hline
\addlinespace[0.15em]

VTimeLLM &
This is a video with \{num\_frames\} frames: \texttt{<video>} \newline
During which frames can we see the impact event ``\{query\}'' happening in the video? If the event occurs multiple separate times, return all intervals, not just the most salient one. Localize only the applied moment of impact, not the approach or aftermath. Return all intervals in chronological order. Answer only in the format: From s1 to e1; From s2 to e2. \\

\addlinespace[0.4em]
TRACE &
\texttt{<video>} \newline 
Localize a series of moments in the video where the impact on a person's body described by ``\{query\}'' is physically applied, and output the start and end timestamps for each matching moment. The query may match multiple separate impact moments in the same video. Return all matching intervals, not just the most salient one. Localize only the applied moment of impact, not the approach or aftermath. \\

\addlinespace[0.4em]
Grounded-Video-LLM &
\texttt{<video>} \texttt{<grounding>} \newline
Give you a textual query: ``\{query\}''. The described content may occur multiple times in the video, where a physical impact is applied to a person's body. When do all the described impact moments occur in the video? Localize only the moment when the impact is physically applied to the person, not the approach or aftermath. Please return all start and end timestamps in chronological order. \\

\addlinespace[0.4em]
UniTime &
This is a sequence interleaved with timestamps and frames. \newline
Your task is to identify the temporal window(s) (start and end timestamps) when the physical impact on a person's body described by the given query is applied. \newline
The query may match multiple separate impact moments in the video. \newline
Localize only the applied moment of impact, not the approach or aftermath. \newline
Query: \{query\} \newline
Answer: \\

\addlinespace[0.4em]
TimeLens &
You are given a video with multiple frames. The numbers before each video frame indicate its sampling timestamp (in seconds). Please find the physical impact on a person's body described by the sentence ``\{query\}'', determining its starting and ending times. The query may match multiple separate impact moments in the same video. Localize only the moment of impact, not the approach or aftermath. The format should be: ``The event happens in <start time 1> - <end time 1> seconds; <start time 2> - <end time 2> seconds; ...''. \\

\addlinespace[0.15em]
\hline
\addlinespace[0.4em]

Qwen3-VL / InternVL-3.5 &
Identify one or more time intervals in this video where the impact on a person's body described by ``\{query\}'' is physically applied. The query may match multiple separate impact moments in the same video. Return all matching intervals, not just the most salient one. Localize only the moment when the impact is applied to the person, not the approach or aftermath. Return only the list of relevant intervals in seconds as \texttt{[[start\_1, end\_1], ..., [start\_n, end\_n]]}, for example \texttt{[[1.040, 1.375], [4.833, 5.125], [7.208, 7.417]]}. If there is only one interval, return it as a nested list \texttt{[[start, end]]}. Do not return any explanation or extra text. \\

\addlinespace[0.4em]
Gemini 3.1 Pro &
Localize one or more time intervals in this video where the impact on a person's body described by ``\{query\}'' is physically applied. The query may match multiple separate impact moments in the same video. Return all matching intervals, not just the most salient one. Localize only the moment when the impact is applied to the person, not the approach or aftermath. Return only the list of relevant intervals in the format \texttt{[[MM:SS.SSS, MM:SS.SSS], ..., [MM:SS.SSS, MM:SS.SSS]]}. For example \texttt{[[00:01.040, 00:01.375], [00:04.833, 00:05.125], [00:07.208, 00:07.417]]}. If there is only one interval, return it as a nested list \texttt{[[MM:SS.SSS, MM:SS.SSS]]}. Do not return any explanation or extra text. \\

\addlinespace[0.4em]
GPT-4o / GPT-5.4 &
You are given multiple frames from a video. Each frame is preceded by its timestamp, e.g., ``Frame at 2.375s:''. Please find one or more time intervals where the impact on a person's body described by the sentence ``\{query\}'' is physically applied, determining their starting and ending times. The query may match multiple separate impact moments in the same video. Return all matching intervals, not just the most salient one. Localize only the moment when the impact is actually applied to the person's body, not the approach or aftermath. Answer in the format ``The event happens in x1 - y1 seconds; x2 - y2 seconds; ...'', where each xi is a start time, and each yi is an end time, with xi $<$ yi. Return all matching moments in chronological order. \\

\addlinespace[0.15em]
\hline
\end{tabularx}

\label{tab:generation_model_prompts}
\end{table*}

For generation-based baselines, the input video was provided either as uniformly down-sampled frames or as a raw video file, depending on the model interface. For frame-based inputs, the recommended frame sampling setting of each model was followed when available. Specifically, VTimeLLM, TRACE, and Grounded-Video-LLM were given $100$, $128$, and $96$ sampled frames, respectively, while the remaining models received up to $192$ frames. For videos shorter than the target number of sampled frames, input frames were limited by the video length. Each model was prompted to output one or more temporal intervals corresponding to the queried impact event. Because the baselines follow different input conventions, we adapt the prompt to each model's native interface while keeping the impact-localization instruction consistent. Table~\ref{tab:generation_model_prompts} lists the resulting prompts. We used deterministic settings for all open-source models, but evaluated each closed-source MLLM over five trials and reported the mean with standard deviation.

\subsubsection{Spatial Localization Evaluation}
\label{sec:spatial_localization_evaluation}

\begin{figure*}[t]
\centering
\includegraphics[width=\textwidth]{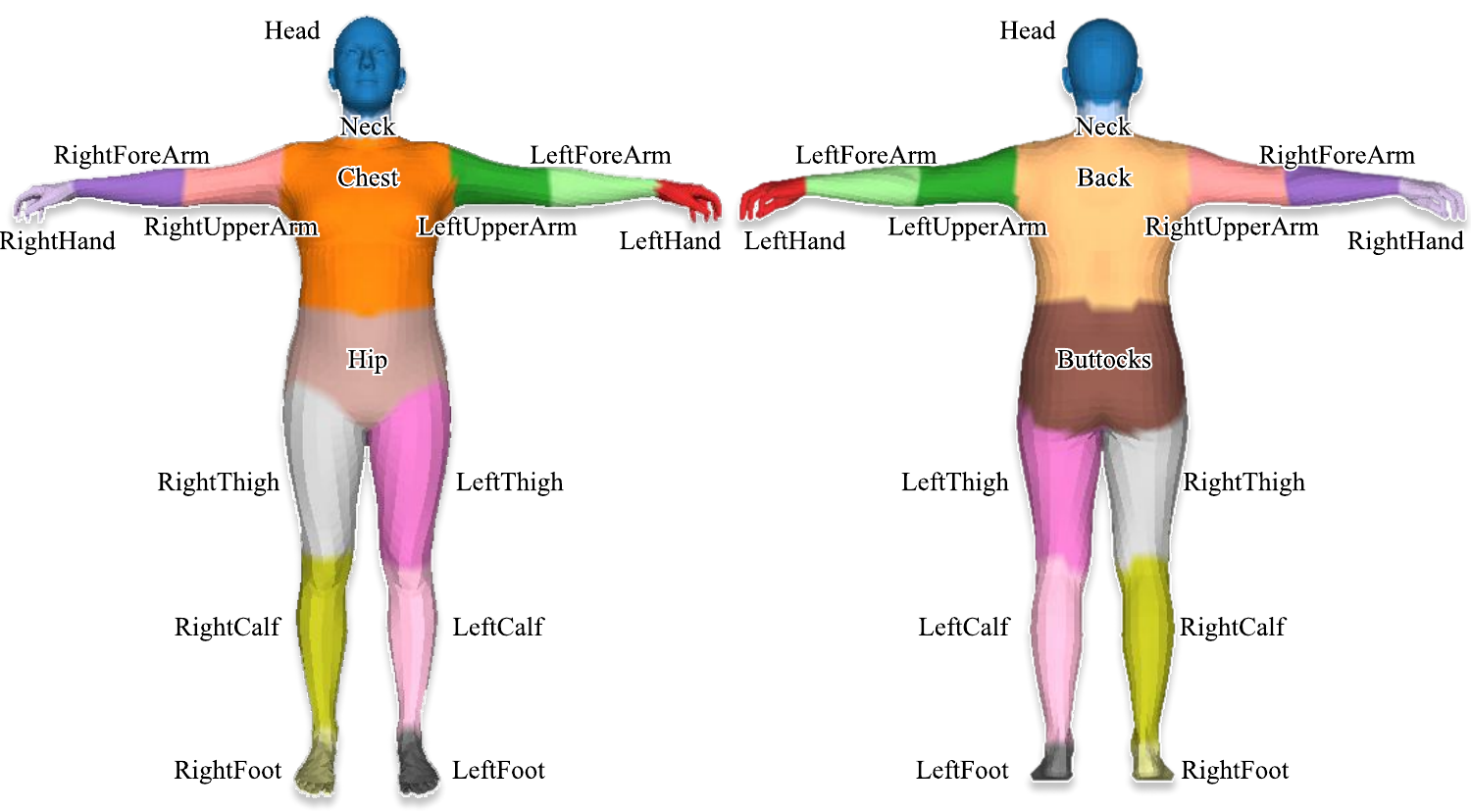}
\caption{The 18 body-part labels for query-guided region selection. Left: front; right: back.}
\label{fig:smplx_part_labels}
\end{figure*}

We select BSTRO~\citep{huang2022capturing}, DECO~\citep{tripathi2023deco}, and GraphiContact~\citep{lin2026graphicontact} as baselines. We exclude methods that require auxiliary context beyond human and scene observations, such as object meshes or hand-held object annotations, since such inputs are not consistently available in movie clips. 

Spatial evaluation was conducted on each frame within the annotated temporal intervals of the queried impact event. For human-scene contact baselines, models were fine-tuned from the officially released pretrained weights. BSTRO was initialized from RICH-pretrained weights and fine-tuned with a learning rate of $5\times10^{-5}$. DECO and GraphiContact were initialized from DAMON-pretrained weights and fine-tuned with learning rates of $5\times10^{-6}$ and $8\times10^{-3}$, respectively. DECO was trained in its no-context setting for image-only evaluation, while GraphiContact was fine-tuned without its semantic scene and body-part segmentation auxiliary losses. All three models were trained for $100$ epochs with checkpoint selection on the validation split. For baselines whose released checkpoints predict only on the SMPL topology, we converted our SMPL-X annotations to SMPL before training and evaluation. 

For Cosmos-Reason2, spatial localization was reformulated as query-guided region selection. The candidate regions were defined by extending the body-part segmentation of \citet{chen2023detecting} with the neck region from the official SMPL-X body segmentation, resulting in $18$ body-part labels. Figure~\ref{fig:smplx_part_labels} shows these part labels as color-coded regions on the SMPL-X mesh. Since this reformulation predicts body-part labels rather than vertices, we further report label-level F1 scores, with the results summarized in Figure~\ref{fig:cosmos_region_f1}.

Both the $2$B and $8$B variants were fine-tuned using the HuggingFace implementation with LoRA. The LoRA setup followed rank $r=32$, $\alpha=32$, dropout $0.05$, a learning rate of $5\times10^{-5}$, and $50$ warmup steps. Both variants were trained for $3$ epochs, and all evaluations used deterministic settings. The input-output format is provided below:

\begin{table*}[!h]
\centering
\setlength{\tabcolsep}{8pt}
\renewcommand{\arraystretch}{1.15}
\begin{tabularx}{\linewidth}{>{\raggedright\arraybackslash\setlength{\parindent}{0pt}}X}
\hline
\rowcolor{black!18}
Prompt for Query-Guided Region Selection \\
\hline
\addlinespace[0.5em]
You are observing a movie frame. The impact query is ``\{query\}''. Select all body regions physically affected by the impact in this frame. \\
\addlinespace[0.5em]
The possible body regions are: head, neck, chest, back, left upper arm, left forearm, left hand, right upper arm, right forearm, right hand, butt, hip, left thigh, left calf, left foot, right thigh, right calf, right foot. \\
\addlinespace[0.5em]
Answer with only the selected body-region labels, separated by commas. \\
\addlinespace[0.3em]
\hline
\end{tabularx}
\label{tab:cosmos_region_prompt}
\end{table*}

\begin{table*}[!h]
\centering
\setlength{\tabcolsep}{8pt}
\renewcommand{\arraystretch}{1.15}
\begin{tabularx}{\linewidth}{>{\raggedright\arraybackslash\setlength{\parindent}{0pt}}X}
\hline
\rowcolor{black!18}
Answer Format \\
\hline
\addlinespace[0.5em]
The affected body regions are \{list of body-region labels\}. \\
\addlinespace[0.3em]
\hline
\end{tabularx}
\label{tab:cosmos_region_answer_format}
\end{table*}

\begin{figure*}[t]
\centering
\includegraphics[width=\textwidth]{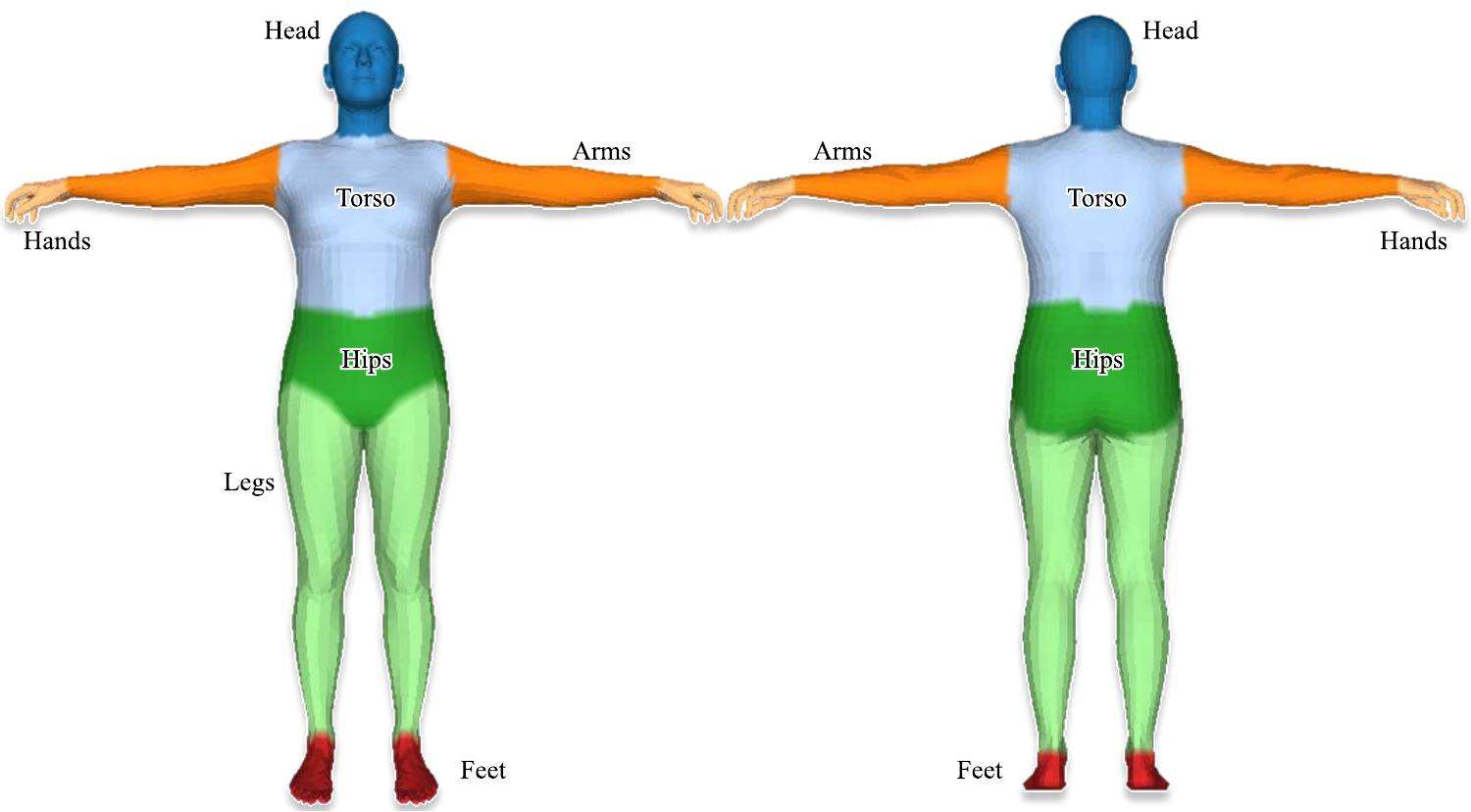}
\caption{The 7 body-region groups for per-region evaluation. Left: front; right: back.}
\label{fig:eval_part_labels}
\end{figure*}

To ensure consistency across all baselines, precision, recall, F1, and geodesic error were computed in the same SMPL space, with the results reported in Table~\ref{tab:human_scene_contact}. We further report F1 scores separately for each body region after merging the body-part labels into $7$ regions. For each region, F1 was computed only on frames where that region was affected in the ground truth, using the vertices belonging to that region. The final score was averaged over those selected frames. Figure~\ref{fig:eval_part_labels} shows the region groups.

\subsubsection{Ablation Settings}

\paragraph{Effect of temporal sampling interval.}
We first ablated the temporal sampling interval for retrieval-based baselines. Moment-DETR and QD-DETR were trained on the HAPACT training split and evaluated on the test split under different sampling intervals, as shown in Table~\ref{tab:ablation_clip_length}. For intervals of $0.5$s or longer, models used both CLIP and SlowFast features, while intervals shorter than $0.5$s used only CLIP features. Overall, denser sampling improved temporal localization performance, with the $0.05$s interval achieving the competitive results. Since human-centric impact localization requires fine-grained, frame-level temporal supervision, we adopted the $0.05$s sampling interval for the main retrieval-based evaluations.

\begin{table}[p]
\centering
\caption{\textbf{Effect of Temporal Sampling Interval.} Moment-DETR and QD-DETR are trained on the HAPACT training split and evaluated on the test split under different temporal sampling intervals.}
\label{tab:ablation_clip_length}
\resizebox{\textwidth}{!}{
\begin{tabular}{@{}lc rr rrrr rrr@{}}
\toprule[1.5pt]
\multirow{2}{*}[-0.5ex]{Method} & \multirowcell{2}[-0.5ex]{Sample \\ Int.} & \multicolumn{2}{c}{} & \multicolumn{4}{c}{mAP} & \multicolumn{3}{c}{mR@k} \\
\cmidrule(lr){5-8} \cmidrule(lr){9-11}
 & & mAP@0.3 & R1@0.3 & G-mAP & @1\_tgt & @2\_tgt & @3+tgt & @1 & @2 & @3 \\
 \midrule
\multirow{5}{*}{Moment-DETR} 
 & 2.0s & \cellcolor[HTML]{C7E9C0}15.65 & \cellcolor[HTML]{E5F5E0}12.70 & \cellcolor[HTML]{A1D99B}16.54 & \cellcolor[HTML]{A1D99B}17.44 & \cellcolor[HTML]{A1D99B}16.66 & \cellcolor[HTML]{F7FCF5}6.99 & \cellcolor[HTML]{C7E9C0}13.67 & \cellcolor[HTML]{C7E9C0}14.91 & \cellcolor[HTML]{E5F5E0}12.87 \\
 & 1.0s & \cellcolor[HTML]{E5F5E0}10.29 & \cellcolor[HTML]{F7FCF5}9.93 & \cellcolor[HTML]{E5F5E0}11.89 & \cellcolor[HTML]{C7E9C0}13.07 & \cellcolor[HTML]{E5F5E0}11.31 & \cellcolor[HTML]{F7FCF5}8.98 & \cellcolor[HTML]{E5F5E0}11.96 & \cellcolor[HTML]{E5F5E0}10.54 & \cellcolor[HTML]{F7FCF5}6.51 \\
 & 0.5s & \cellcolor[HTML]{E5F5E0}11.12 & \cellcolor[HTML]{E5F5E0}12.01 & \cellcolor[HTML]{E5F5E0}11.40 & \cellcolor[HTML]{E5F5E0}12.10 & \cellcolor[HTML]{E5F5E0}11.27 & \cellcolor[HTML]{F7FCF5}8.01 & \cellcolor[HTML]{E5F5E0}12.75 & \cellcolor[HTML]{E5F5E0}11.03 & \cellcolor[HTML]{F7FCF5}8.99 \\
 & 0.1s$^*$ & \cellcolor[HTML]{A1D99B}16.98 & \cellcolor[HTML]{C7E9C0}13.86 & \cellcolor[HTML]{74C476}18.65 & \cellcolor[HTML]{74C476}19.82 & \cellcolor[HTML]{A1D99B}17.30 & \cellcolor[HTML]{E5F5E0}12.22 & \cellcolor[HTML]{C7E9C0}14.60 & \cellcolor[HTML]{A1D99B}16.00 & \cellcolor[HTML]{74C476}19.38 \\
 & 0.05s$^*$ & \cellcolor[HTML]{A1D99B}16.25 & \cellcolor[HTML]{E5F5E0}12.24 & \cellcolor[HTML]{A1D99B}17.70 & \cellcolor[HTML]{A1D99B}17.19 & \cellcolor[HTML]{41AB5D}20.11 & \cellcolor[HTML]{C7E9C0}13.25 & \cellcolor[HTML]{C7E9C0}13.63 & \cellcolor[HTML]{74C476}18.85 & \cellcolor[HTML]{74C476}18.14 \\
\midrule
\multirow{5}{*}{QD-DETR} 
 & 2.0s & \cellcolor[HTML]{A1D99B}16.60 & \cellcolor[HTML]{E5F5E0}11.32 & \cellcolor[HTML]{74C476}19.68 & \cellcolor[HTML]{41AB5D}20.05 & \cellcolor[HTML]{41AB5D}21.14 & \cellcolor[HTML]{E5F5E0}10.31 & \cellcolor[HTML]{C7E9C0}14.60 & \cellcolor[HTML]{74C476}18.97 & \cellcolor[HTML]{A1D99B}16.90 \\
 & 1.0s & \cellcolor[HTML]{C7E9C0}13.67 & \cellcolor[HTML]{F7FCF5}9.24 & \cellcolor[HTML]{A1D99B}16.25 & \cellcolor[HTML]{C7E9C0}15.67 & \cellcolor[HTML]{74C476}18.90 & \cellcolor[HTML]{F7FCF5}9.50 & \cellcolor[HTML]{E5F5E0}12.05 & \cellcolor[HTML]{C7E9C0}15.64 & \cellcolor[HTML]{A1D99B}16.90 \\
 & 0.5s & \cellcolor[HTML]{A1D99B}17.51 & \cellcolor[HTML]{C7E9C0}14.09 & \cellcolor[HTML]{41AB5D}20.13 & \cellcolor[HTML]{41AB5D}21.05 & \cellcolor[HTML]{41AB5D}20.21 & \cellcolor[HTML]{E5F5E0}11.91 & \cellcolor[HTML]{A1D99B}17.00 & \cellcolor[HTML]{74C476}18.30 & \cellcolor[HTML]{A1D99B}17.83 \\
 & 0.1s$^*$ & \cellcolor[HTML]{A1D99B}16.02 & \cellcolor[HTML]{E5F5E0}12.93 & \cellcolor[HTML]{74C476}18.84 & \cellcolor[HTML]{74C476}18.94 & \cellcolor[HTML]{41AB5D}20.63 & \cellcolor[HTML]{F7FCF5}8.74 & \cellcolor[HTML]{C7E9C0}15.43 & \cellcolor[HTML]{74C476}19.40 & \cellcolor[HTML]{C7E9C0}15.66 \\
 & 0.05s$^*$ & \cellcolor[HTML]{41AB5D}20.06 & \cellcolor[HTML]{A1D99B}16.63 & \cellcolor[HTML]{41AB5D}21.36 & \cellcolor[HTML]{41AB5D}22.36 & \cellcolor[HTML]{41AB5D}21.01 & \cellcolor[HTML]{E5F5E0}12.83 & \cellcolor[HTML]{74C476}18.20 & \cellcolor[HTML]{41AB5D}20.67 & \cellcolor[HTML]{74C476}18.29 \\
\bottomrule[1.5pt] 
\multicolumn{11}{@{}l}{\scriptsize{$^*$ Models trained with only CLIP features (intervals $< 0.5s$ are too short to extract SlowFast features).}} \\
\end{tabular}
}
\end{table}

\begin{table}[p]
\centering
\caption{\textbf{Effect of Pre-training and Fine-tuning.} 
Retrieval-based models implemented in the Lighthouse~\citep{nishimura2024lighthouse} codebase are evaluated under three settings: QVHighlights pre-trained checkpoints without fine-tuning (\emph{Zero-shot}), training from scratch on the training split (\emph{HAPACT}), and fine-tuning the QVHighlights checkpoints on the training split (\emph{Fine-tune}). All models use CLIP features with a temporal sampling interval of $0.05$ seconds and are evaluated on the HAPACT test split.}
\label{tab:ablation_pretraining_finetuning}
\resizebox{\textwidth}{!}{
\begin{tabular}{@{}lc rr rrrr rrr@{}}
\toprule[1.5pt]
\multirow{2}{*}[-0.5ex]{Method} & \multirow{2}{*}[-0.5ex]{Setting} 
& \multicolumn{2}{c}{} & \multicolumn{4}{c}{mAP} & \multicolumn{3}{c}{mR@k} \\
\cmidrule(lr){5-8} \cmidrule(lr){9-11}
 & & mAP@0.3 & R1@0.3 & G-mAP & @1\_tgt & @2\_tgt & @3+tgt & @1 & @2 & @3 \\
 \midrule
\multirow{3}{*}{Moment-DETR}
 & \emph{Zero-shot} & 8.05 & 5.77 & 10.04 & 11.69 & 7.07 & 5.56 & 6.93 & 5.88 & 7.29 \\
 & \emph{HAPACT} & 16.25 & 12.24 & 17.70 & 17.19 & 20.11 & 13.25 & 13.63 & 18.85 & 18.14 \\
 & \cellcolor[HTML]{E0E0E0}\emph{\textbf{Fine-tune}} & \cellcolor[HTML]{E0E0E0}18.16 & \cellcolor[HTML]{E0E0E0}13.63 & \cellcolor[HTML]{E0E0E0}20.39 & \cellcolor[HTML]{E0E0E0}21.94 & \cellcolor[HTML]{E0E0E0}19.11 & \cellcolor[HTML]{E0E0E0}12.92 & \cellcolor[HTML]{E0E0E0}15.57 & \cellcolor[HTML]{E0E0E0}16.61 & \cellcolor[HTML]{E0E0E0}16.59 \\
\midrule
\multirow{3}{*}{QD-DETR}
 & \emph{Zero-shot} & 9.55 & 6.00 & 11.98 & 13.57 & 9.50 & 7.81 & 8.22 & 8.12 & 8.53 \\
 & \emph{HAPACT} & 20.06 & 16.63 & 21.36 & 22.36 & 21.01 & 12.83 & 18.20 & 20.67 & 18.29 \\
 & \cellcolor[HTML]{E0E0E0}\emph{\textbf{Fine-tune}} & \cellcolor[HTML]{E0E0E0}21.27 & \cellcolor[HTML]{E0E0E0}16.17 & \cellcolor[HTML]{E0E0E0}22.06 & \cellcolor[HTML]{E0E0E0}23.91 & \cellcolor[HTML]{E0E0E0}19.98 & \cellcolor[HTML]{E0E0E0}15.39 & \cellcolor[HTML]{E0E0E0}16.86 & \cellcolor[HTML]{E0E0E0}18.67 & \cellcolor[HTML]{E0E0E0}19.69 \\
\midrule
\multirow{3}{*}{EaTR}
 & \emph{Zero-shot} & 9.49 & 6.24 & 12.28 & 12.85 & 11.78 & 6.29 & 8.13 & 9.64 & 11.01 \\
 & \emph{HAPACT} & 16.65 & 13.63 & 19.62 & 20.47 & 18.38 & 11.53 & 16.21 & 15.58 & 16.90 \\
 & \cellcolor[HTML]{E0E0E0}\emph{\textbf{Fine-tune}} & \cellcolor[HTML]{E0E0E0}21.69 & \cellcolor[HTML]{E0E0E0}16.17 & \cellcolor[HTML]{E0E0E0}22.54 & \cellcolor[HTML]{E0E0E0}23.09 & \cellcolor[HTML]{E0E0E0}22.58 & \cellcolor[HTML]{E0E0E0}13.92 & \cellcolor[HTML]{E0E0E0}16.95 & \cellcolor[HTML]{E0E0E0}19.21 & \cellcolor[HTML]{E0E0E0}20.62 \\
\midrule
\multirow{3}{*}{TR-DETR}
 & \emph{Zero-shot} & 10.62 & 7.39 & 14.00 & 14.97 & 12.81 & 6.99 & 10.21 & 9.64 & 11.01 \\
 & \emph{HAPACT} & 13.62 & 10.16 & 14.87 & 15.78 & 13.91 & 9.36 & 11.46 & 11.76 & 12.40 \\
 & \cellcolor[HTML]{E0E0E0}\emph{\textbf{Fine-tune}} & \cellcolor[HTML]{E0E0E0}20.86 & \cellcolor[HTML]{E0E0E0}16.17 & \cellcolor[HTML]{E0E0E0}22.20 & \cellcolor[HTML]{E0E0E0}23.86 & \cellcolor[HTML]{E0E0E0}20.36 & \cellcolor[HTML]{E0E0E0}14.62 & \cellcolor[HTML]{E0E0E0}17.69 & \cellcolor[HTML]{E0E0E0}17.94 & \cellcolor[HTML]{E0E0E0}21.39 \\
\midrule
\multirow{3}{*}{CG-DETR}
 & \emph{Zero-shot} & 7.96 & 4.62 & 11.09 & 13.05 & 8.27 & 5.08 & 7.11 & 6.06 & 7.29 \\
 & \emph{HAPACT} & 16.10 & 11.09 & 17.38 & 18.32 & 16.09 & 9.42 & 12.65 & 13.76 & 15.50 \\
 & \cellcolor[HTML]{E0E0E0}\emph{\textbf{Fine-tune}} & \cellcolor[HTML]{E0E0E0}20.57 & \cellcolor[HTML]{E0E0E0}15.47 & \cellcolor[HTML]{E0E0E0}22.55 & \cellcolor[HTML]{E0E0E0}23.96 & \cellcolor[HTML]{E0E0E0}19.48 & \cellcolor[HTML]{E0E0E0}17.28 & \cellcolor[HTML]{E0E0E0}17.92 & \cellcolor[HTML]{E0E0E0}21.52 & \cellcolor[HTML]{E0E0E0}22.63 \\
\midrule
\multirow{3}{*}{UVCOM}
 & \emph{Zero-shot} & 10.20 & 7.39 & 13.65 & 15.56 & 10.71 & 5.98 & 10.67 & 8.12 & 8.22 \\
 & \emph{HAPACT} & 18.01 & 16.17 & 19.77 & 22.25 & 15.55 & 9.46 & 18.15 & 18.24 & 19.54 \\
 & \cellcolor[HTML]{E0E0E0}\emph{\textbf{Fine-tune}} & \cellcolor[HTML]{E0E0E0}21.22 & \cellcolor[HTML]{E0E0E0}17.32 & \cellcolor[HTML]{E0E0E0}23.58 & \cellcolor[HTML]{E0E0E0}24.28 & \cellcolor[HTML]{E0E0E0}22.96 & \cellcolor[HTML]{E0E0E0}16.53 & \cellcolor[HTML]{E0E0E0}19.26 & \cellcolor[HTML]{E0E0E0}21.76 & \cellcolor[HTML]{E0E0E0}24.65 \\
\bottomrule[1.5pt]
\end{tabular}
}
\end{table}

\begin{table}[p]
\centering
\caption{\textbf{Effect of Input Modality in the MLLM Baseline.}
Text-only (\emph{T}), image-only (\emph{I}), and text-image (\emph{T+I}) inputs are compared under the same query-guided region selection setting. The table reports overall localization performance and per-region F1 over seven coarse body-region groups.}
\label{tab:ablation_modality}
\resizebox{\textwidth}{!}{
\begin{tabular}{@{}l crrrr rrrrrrr@{}}
\toprule[1.5pt]
\multirow{2}{*}{Method} & \multirow{2}{*}{Modality} & \multirow{2}{*}{F1 $\uparrow$} & \multirow{2}{*}{Precision $\uparrow$} & \multirow{2}{*}{Recall $\uparrow$} & \multirow{2}{*}{geo. (cm) $\downarrow$} & \multicolumn{7}{c}{F1 Scores per Body-part $\uparrow$} \\
\cmidrule(lr){7-13}
 & & & & & & Head & Arms & Torso & Hips & Legs & Hands & Feet \\
 \midrule
\multirow{3}{*}{Cosmos-Reason2-8B}& \emph{T} & 15.98 & 14.03 & 39.36 & 37.84 & 20.90 & 13.90 & 19.68 & 24.35 & 7.66 & 12.29 & 8.13\\
& \emph{I} & 16.46 & 14.48 & 36.02 & 39.65 & 19.83 & 15.99 & 19.26 & 26.47 & 10.76 & 17.79 & 7.60\\
& \cellcolor[HTML]{E0E0E0}\emph{T + I} & \cellcolor[HTML]{E0E0E0}18.59 & \cellcolor[HTML]{E0E0E0}16.17 & \cellcolor[HTML]{E0E0E0}44.11 & \cellcolor[HTML]{E0E0E0}33.31 & \cellcolor[HTML]{E0E0E0}21.21 & \cellcolor[HTML]{E0E0E0}15.92 & \cellcolor[HTML]{E0E0E0}20.01 & \cellcolor[HTML]{E0E0E0}29.95 & \cellcolor[HTML]{E0E0E0}12.59 & \cellcolor[HTML]{E0E0E0}17.87 & \cellcolor[HTML]{E0E0E0}8.71\\
\bottomrule[1.5pt]
\end{tabular}
}
\end{table}

\paragraph{Effect of pre-training and fine-tuning.}
We further examined whether retrieval-based temporal models benefit more from task-specific training alone or from adapting pretrained moment-retrieval checkpoints. As shown in Table~\ref{tab:ablation_pretraining_finetuning}, models initialized from QVHighlights and fine-tuned on HAPACT generally achieved stronger performance than models trained from scratch, indicating that prior moment-retrieval supervision provides a useful initialization for human-impact localization. Based on this result, the main retrieval-based temporal baselines were initialized from the pretrained checkpoints then fine-tuned on HAPACT.

\paragraph{Effect of modalities in the MLLM baseline.}
For spatial impact localization, we ablated the input modalities of the MLLM baseline by comparing text-only, image-only, and text-image settings, as shown in Table~\ref{tab:ablation_modality}. The text-image setting achieved the strongest performance, suggesting that impact queries and visual evidence provide complementary cues. Accordingly, the main MLLM baseline incorporates both modalities.

\subsection{More Benchmark Results and Analyses}
\label{sec:additional_benchmark_results}

\subsubsection{Analysis of Temporal Bottlenecks in VMR Models}
\label{sec:temporal_bottleneck_analysis}

Results in Table~\ref{tab:direct_vmr_models} and Table~\ref{tab:gen_vmr_models} suggest different performance patterns across model families. To better understand the temporal bottlenecks of each family, we conduct subgroup analyses by impact length and ground-truth moment count per query. We evaluate a strong representative from each model family: Sim-DETR, UniTime, Qwen3-VL-8B, and GPT-5.4, and summarize their subgroup performance in Table~\ref{tab:temporal_bottlenecks}. Sim-DETR and Qwen3-VL-8B are strongly affected by shorter impacts, whereas GPT-5.4 shows a smaller relative decline. Multiple-moment queries lead to substantially larger drops for Qwen3-VL-8B and GPT-5.4 than for Sim-DETR. UniTime remains low across both analyses. These results show different sensitivities to impact length and the number of moments across the evaluated VMR models, while other temporal bottlenecks may also remain.

\subsubsection{HSC Baseline Performance under Different Training Settings}
\label{sec:additional_hsc_results}

To further evaluate whether HSC architectures can learn impact localization from task-specific supervision alone, we additionally train DECO and GraphiContact from scratch on HAPACT. BSTRO is excluded because its released implementation requires initialization weights that are not publicly available. As shown in Table~\ref{tab:hsc_scratch}, training from scratch does not consistently improve F1 score over fine-tuning across the evaluated models. The limited transfer of HSC models therefore cannot be attributed solely to their original pretraining and may also reflect limitations in their architectures or learning formulations for impact localization.

\subsubsection{Effect of Region-Label Imbalance on Spatial Performance}
\label{sec:region_imbalance_analysis}

Figure~\ref{fig:cosmos_region_f1} shows that our MLLM baseline performs unevenly across body regions, with higher F1 scores on the head, back, and buttocks. The impact occurrence distribution in Figure~\ref{fig:heatmap} is also concentrated in several of these regions across all splits, motivating us to examine whether region-label imbalance contributes to this performance pattern. 

We therefore construct a region-balanced HAPACT training set. Specifically, the original training set is minimally pruned to balance occurrence counts across the 18 body-region labels. After pruning, the occurrence count of each label differs from that of the rarest label by at most two. We additionally construct a randomly down-sampled set with the same number of frames as a control. 

The baseline is then trained separately on each training set and evaluated on the original test set, with the results reported in Table~\ref{tab:region_imbalance}. The region-balanced set improves mean region F1 while reducing both the Region Gap and Across-Region SD. By contrast, random down-sampling does not reduce the gap. These results suggest that the original region-label imbalance may have partly contributed to the baseline's lower overall performance, but does not fully explain the disparity.

\begin{table}[p]
\centering
\small
\caption{\textbf{Effects of Impact Duration and Moment Count on Temporal Localization.}
Values in parentheses denote relative changes from the reference group, with darker red indicating larger decreases. For GPT-5.4, results are reported as mean $\pm$ standard deviation over five runs, with relative changes computed from the mean. \textbf{Bold} indicates the smallest decrease for each comparison.}
\label{tab:temporal_bottlenecks}

\resizebox{\textwidth}{!}{
\begin{tabular}{lrrrrr}
\multicolumn{4}{l}{\color{gray}\textit{(a) Impact Duration}} &
\multicolumn{2}{l}{\color{gray}\textit{(b) Moment Count}} \\[0.15em]
\toprule[1.5pt]
& \multicolumn{3}{c}{mAP@0.3 (\%)} & \multicolumn{2}{c}{mAP@0.3 (\%)} \\
\cmidrule(lr){2-4} \cmidrule(lr){5-6}
Method & \multicolumn{1}{c}{Long (7+f, ref.)} & \multicolumn{1}{c}{Middle (3--6f)} & \multicolumn{1}{c}{Short (1--2f)} & \multicolumn{1}{c}{Single (1, ref.)} & \multicolumn{1}{c}{Multiple (2+)} \\
 \midrule
Sim-DETR & 40.22 & \cellcolor{red!19}24.71 (-38.6\%) & \cellcolor{red!37}10.17 (-74.7\%) & 26.58 & \cellcolor{red!4}24.23 \textbf{(-8.8\%)} \\
UniTime & 27.31 & \cellcolor{red!48}1.04 (-96.2\%) & \cellcolor{red!50}0.00 (-100.0\%) & 16.42 & \cellcolor{red!32}5.90 (-64.1\%) \\
Qwen3-VL-8B & 46.92 & \cellcolor{red!21}26.81 (-42.9\%) & \cellcolor{red!47}3.22 (-93.1\%) & 28.31 & \cellcolor{red!23}15.04 (-46.9\%) \\
GPT-5.4 & 51.32 {\scriptsize$\pm$ 1.38} & \cellcolor{red!7}44.60 {\scriptsize$\pm$ 1.76} \textbf{(-13.1\%)} & \cellcolor{red!25}25.72 {\scriptsize$\pm$ 1.85} \textbf{(-49.9\%)} & 53.90 {\scriptsize$\pm$ 1.38} & \cellcolor{red!28}23.58 {\scriptsize$\pm$ 0.39} (-56.3\%) \\
\bottomrule[1.5pt]
\end{tabular}
}
\end{table}

\begin{table}[p]
\centering
\small
\caption{\textbf{HSC Baseline Performance under Different Training Settings.}
\emph{Zero-shot} denotes evaluation using the released DAMON checkpoints, \emph{HAPACT} denotes training from scratch on HAPACT, and \emph{Fine-tune} denotes fine-tuning the DAMON checkpoints on HAPACT.}
\label{tab:hsc_scratch}
\begin{tabular}{lcrrrr}
\toprule[1.5pt]
Method & Setting & F1 (\%) $\uparrow$ & Precision (\%) $\uparrow$ & Recall (\%) $\uparrow$ & geo. (cm) $\downarrow$ \\
 \midrule
\multirow{3}{*}{DECO}
 & \emph{Zero-shot} & 8.28 & 9.87 & 11.84 & 83.84 \\
 & \cellcolor[HTML]{F0F0F0}\emph{HAPACT} & \cellcolor[HTML]{F0F0F0}5.05 & \cellcolor[HTML]{F0F0F0}68.92 & \cellcolor[HTML]{F0F0F0}6.05 & \cellcolor[HTML]{F0F0F0}55.75 \\
 & \emph{Fine-tune} & 5.41 & 68.16 & 6.18 & 50.34 \\
\midrule
\multirow{3}{*}{GraphiContact}
 & \emph{Zero-shot} & 7.88 & 8.85 & 10.23 & 87.42 \\
 & \cellcolor[HTML]{F0F0F0}\emph{HAPACT} & \cellcolor[HTML]{F0F0F0}7.74 & \cellcolor[HTML]{F0F0F0}8.71 & \cellcolor[HTML]{F0F0F0}10.10 & \cellcolor[HTML]{F0F0F0}88.11 \\
 & \emph{Fine-tune} & 7.34 & 30.12 & 14.18 & 50.02 \\
\bottomrule[1.5pt]
\end{tabular}
\end{table}

\begin{table}[p]
\centering
\small
\caption{\textbf{Effect of Region-Label Imbalance on the Spatial Performance of the MLLM baseline.}
Mean Region F1 is the mean F1 across all 18 body-region labels. Head/Torso comprises the head, neck, chest, back, hip, and buttocks, and the remaining 12 labels form Limb. Region Gap is the difference between the mean F1 scores of the Head/Torso and Limb groups. Across-Region SD is the standard deviation of F1 scores across all regions. Lower disparity values indicate more uniform performance. \textbf{Bold} indicates the best result.}
\label{tab:region_imbalance}

\setlength{\tabcolsep}{4pt}
\begin{tabular}{@{}l rrr rr@{}}
\toprule[1.5pt]
\multirow{2}{*}{Training Set}
& \multicolumn{3}{c}{Region-level F1 (\%) $\uparrow$}
& \multicolumn{2}{c}{Disparity $\downarrow$} \\
\cmidrule(lr){2-4} \cmidrule(lr){5-6}
& Mean Region & Head/Torso & Limb & Region Gap & Across-Region SD \\
 \midrule
Original (HAPACT) & 36.57 & 49.54 & 30.09 & 19.45 & 12.63 \\
Random Down-Sampling & 37.25 & \textbf{50.37} & 30.68 & 19.69 & 13.54 \\
Region-Balanced Down-Sampling & \textbf{38.04} & 49.05 & \textbf{32.54} & \textbf{16.51} & \textbf{11.51} \\
\bottomrule[1.5pt]
\end{tabular}
\end{table}

\clearpage
\newpage

\subsection{Dataset Examples}

This section presents representative examples from the HAPACT training and validation sets to illustrate the structure of each video-query instance and its annotations. Each example consists of a movie clip, a textual query, one or more temporal intervals, and frame-level on-body impact labels. An annotation visualization video for these examples is included in the supplementary material.

\begin{figure*}[!h]
\centering
\includegraphics[width=\linewidth]{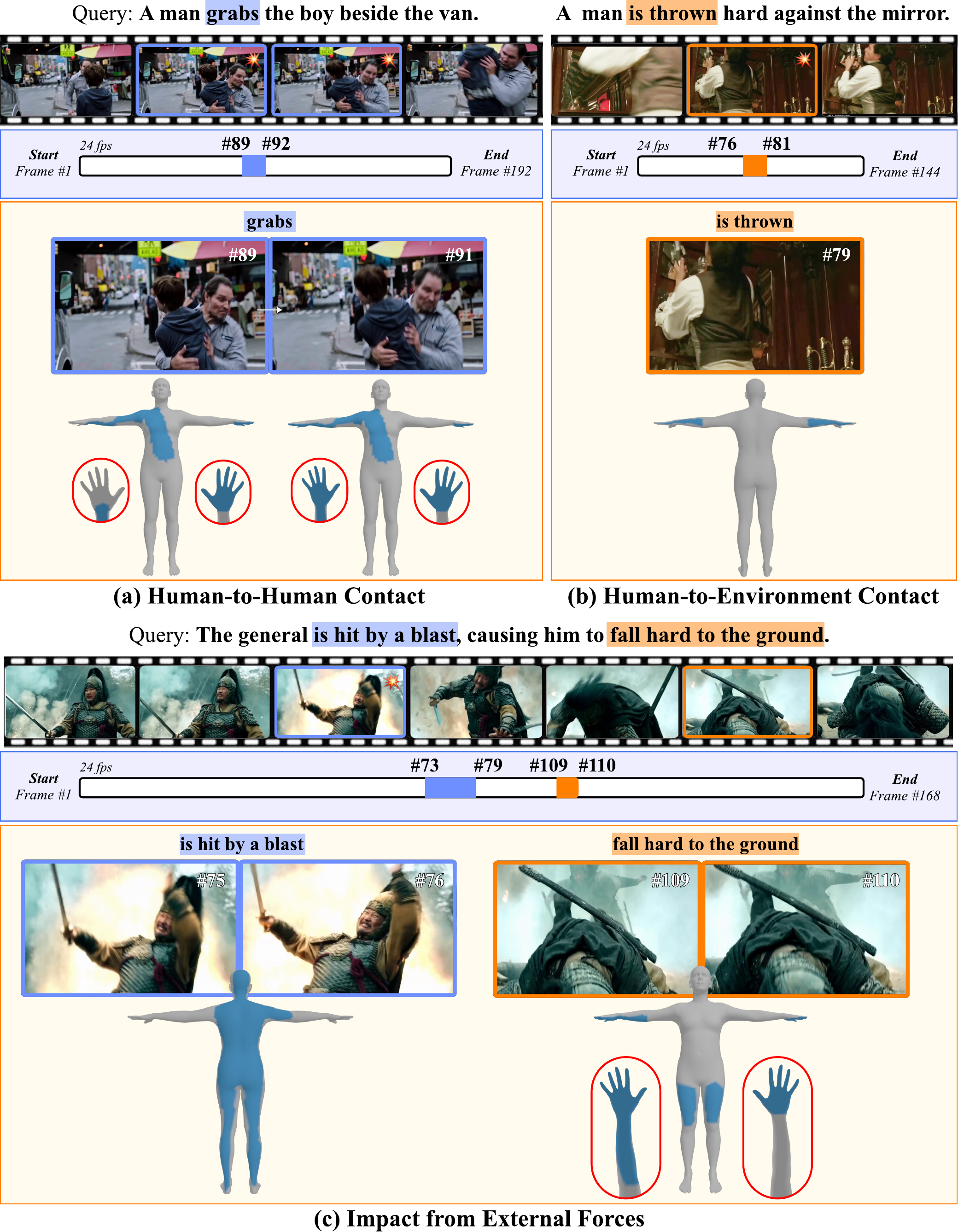}
\caption{\textbf{Examples of HAPACT.} The examples show three representative impact categories: (a) human-to-human contact (e.g., grabbing), (b) human-to-environment contact (e.g., being thrown against a surface), and (c) impact from external forces (e.g., blast explosions).}
\label{fig:sample}
\end{figure*}

\clearpage
\newpage
\subsection{Limitations}
\label{sec:limitations}

\begin{figure*}[t]
\centering
\includegraphics[width=0.86\textwidth]{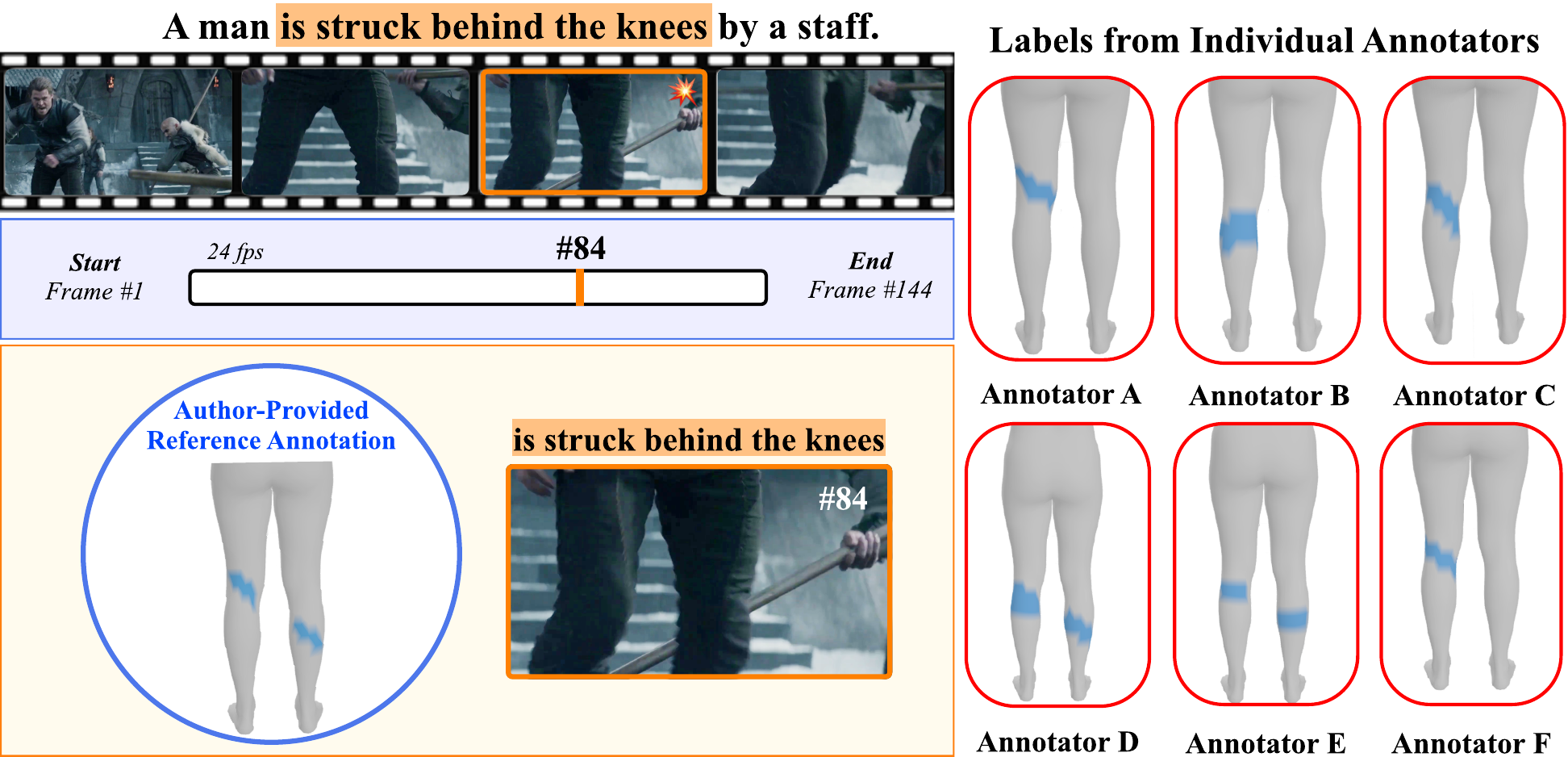}
\caption{\textbf{Annotation subjectivity in spatial localization.}
The reference annotation and individual annotator labels for the same impact frame and query show similar spatial tendencies, while precise vertex labels differ due to occlusion, narrow contact, and annotator interpretation.}
\label{fig:limitation}
\end{figure*}

\paragraph{Annotation Subjectivity.}
Since HAPACT relies on human annotations, its labels may reflect how annotators interpret physical impacts from visual evidence and prior knowledge. This limitation is particularly relevant for spatial localization, especially when the impact is brief and affects a small body region. Figure~\ref{fig:limitation} illustrates this ambiguity. For the query ``A man is struck behind the knees by a staff'' at frame 84, the author-provided annotation and six individual annotator annotations show similar overall tendencies, but differ in the precise annotated regions due to occlusion, the narrow contact area, and annotator-specific interpretations of the impact. Similar challenges have been noted by~\citet{tripathi2023deco}, where contact from images is inherently difficult and often requires human reasoning about occluded body-object relations.

Such ambiguity is a common limitation in human-annotated physical-interaction datasets, particularly at fine granularity. As reported in Section~\ref{sec:quality_control_assessment}, HAPACT achieves Fleiss' $\kappa$ values of $0.53$ at the vertex level and $0.61$ at the body-part level. Related datasets report similar agreement levels: DAMON~\citep{tripathi2023deco} reports $\kappa=0.656$ for manually annotated SMPL contact vertices, and EPIC-Contact~\citep{bansal2026towards} reports $\kappa=0.61$ over $3{,}106$ MANO vertices. Beyond agreement, HAPACT further applies quality-control procedures to maintain quality despite such variability, including annotator pre-screening and meta-annotator review, as detailed in Section~\ref{sec:quality_control_assessment}.

The resulting labels should not be interpreted as perfectly objective physical measurements. Instead, they provide a reliable source for learning the visual and semantic tendencies of human impact localization. Exact vertex selections may remain ambiguous in challenging cases, even when annotators agree on the affected body region. This limitation should be considered for precise vertex-level localization, whereas HAPACT remains well suited to applications where coarse spatial granularity is sufficient, including media-based haptic authoring and haptic experience generation.

\paragraph{Model-Lineage Bias.} 
Gemini 2.5 Pro and GPT-4o were used during candidate generation and filtering, while the final labels were produced through human review and annotation. Although human review reduces direct model influence, lineage-specific biases introduced at these stages may remain. Models from the same lineages may therefore be better aligned with the data distribution, potentially reducing benchmark neutrality for GPT/Gemini-family models.

\paragraph{Limited Real-World Generalizability.}
HAPACT is constructed from movies, primarily motivated by media-based applications where audiovisual and narrative context is important. This focus may lead to domain shifts when applied to real-world videos, and HAPACT does not cover the full range of impacts, such as minor daily bumps, sports impacts, or medical-related collisions. At the same time, HAPACT covers diverse agents, objects, scenes, and impact events, as shown in Figure~\ref{fig:analysis}(f). The data construction pipeline itself is not specific to movies and could be extended to domain-specific videos, although such extensions and validation are left to future work.

\clearpage
%%%%%%%%%%%%%%%%%%%%%%%%%%%%%%%%%%%%%%%%%%%%%%%%%%%%%%%%%%%%

% \newpage
% \input{checklist.tex}

\end{document}